%% file: main.tex
\documentclass[final]{cvpr}

\usepackage{times}
\usepackage{epsfig}
\usepackage{graphicx}
\usepackage{amsmath}
\usepackage{amssymb}
\usepackage[accsupp]{axessibility}
\usepackage{tablefootnote}
\usepackage{booktabs}
\usepackage{overpic}
\usepackage{xcolor}
\usepackage{multirow}
\usepackage{color, colortbl}
\usepackage{subfig}

\ifdefined \GramaCheck
\newcommand{\CheckRmv}[1]{}
\newcommand{\figref}[1]{Figure 1}%
\newcommand{\tabref}[1]{Table 1}%
\newcommand{\secref}[1]{Section 1}
\renewcommand{\eqref}[1]{Equation 1}
\else
\newcommand{\CheckRmv}[1]{#1}
\renewcommand{\eqref}[1]{Eqn.~(\ref{#1})}
\fi
\usepackage[pagebackref=true,letterpaper=true,colorlinks=true,linkcolor=myred,citecolor=mygreen,bookmarks=false]{hyperref}

\definecolor{mygreen}{RGB}{0,150,0}
\definecolor{myred}{RGB}{200,0,0}

\newcommand{\rot}[1]{\rotatebox[]{90}{#1}}
\newcommand{\pub}[1]{$_{\textbf{#1}}$}

\begin{document}

\title{Personalized Image Semantic Segmentation}

\author{Yu Zhang$^{1}$ \qquad Chang-Bin Zhang$^1$ \qquad 
	Peng-Tao Jiang $^1$ \qquad  Ming-Ming Cheng$^1$\thanks{MM Cheng is the corresponding author.} \qquad Feng Mao$^2$ \\
  $^1$TKLNDST, CS, Nankai University \qquad
  $^2$Alibaba Group \\
{\tt\small zhangyuygss@gmail.com \qquad	 pt.jiang@mail.nankai.edu.cn \qquad cmm@nankai.edu.cn }
}

\maketitle
\pagestyle{empty}
\thispagestyle{empty}

\begin{abstract}
Semantic segmentation models trained on public datasets 
have achieved great success in recent years.
However, these models didn't consider the personalization 
issue of segmentation though
it is important in practice.
In this paper, we address the problem of personalized image segmentation.
The objective is to
generate more accurate segmentation results on unlabeled
personalized images 
by investigating the data's personalized traits.
To open up future research in 
this area,
we collect a large dataset containing various users' 
personalized images called PSS (Personalized Semantic Segmentation).
We also survey some recent researches related to this 
problem and
report their performance on our dataset.
Furthermore, by observing the correlation 
among a user's personalized images,
we propose a baseline method that incorporates 
the inter-image context when segmenting certain images.
Extensive experiments show that our method outperforms
the existing methods
on the proposed dataset.
The code and the PSS dataset are available at 
\url{https://mmcheng.net/pss/.}

\end{abstract}

\section{Introduction}
\begin{figure}[t]
	\centering
	\includegraphics[width=0.45\textwidth]{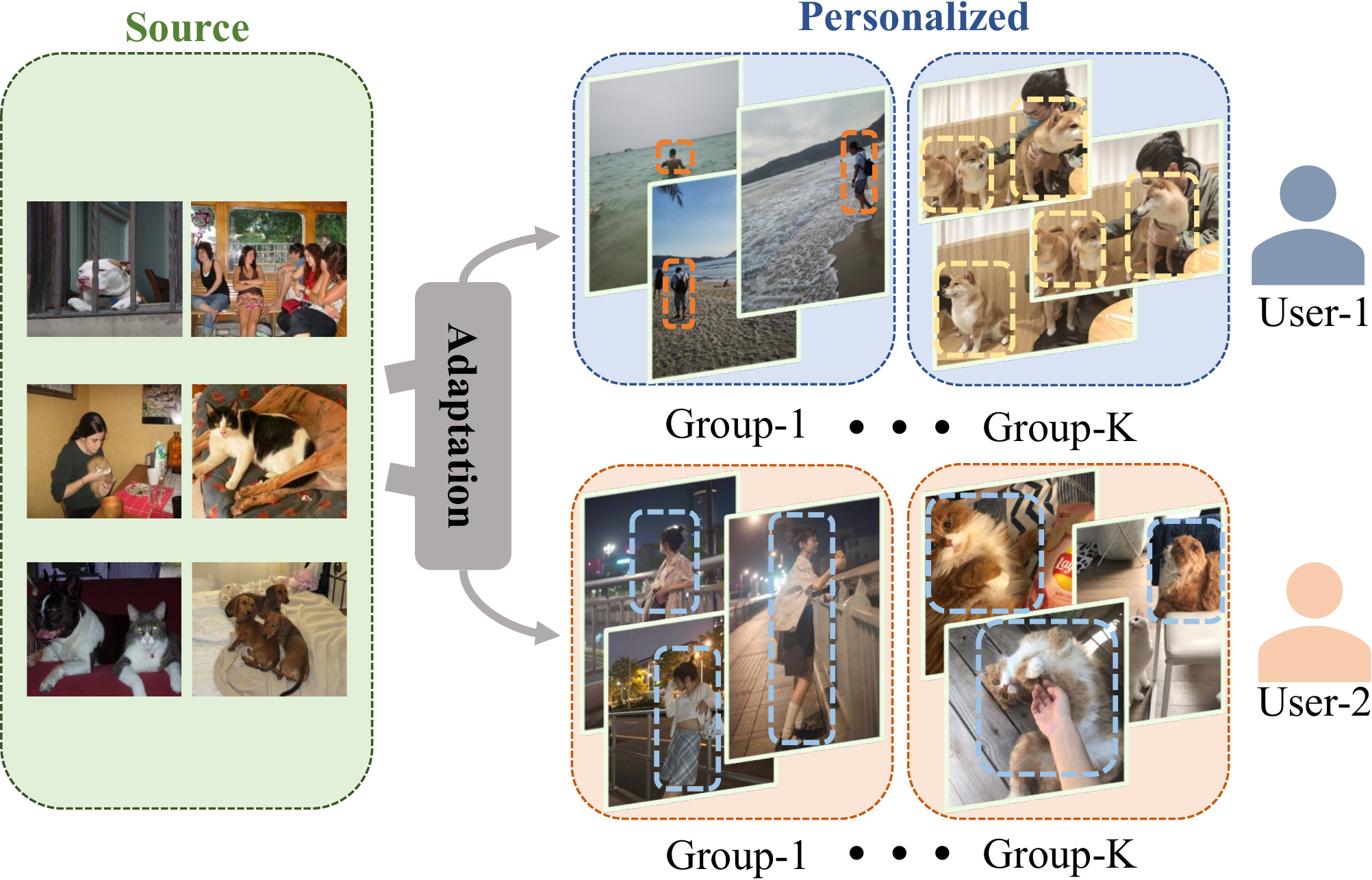}
	\caption{Samples of the proposed personalized image
		dataset. We investigate how to utilize the personalized
		property of users' images when adapting from
		source data. One notable property we can observe is that
		images from the same user are correlated (similar objects, scenes, etc.).
		}
	\label{figure:intro_fig}
\end{figure}

Semantic segmentation is a well-studied task in the computer vision
society. The goal of this task is to assign a
semantic label to each pixel of a given image.
As with other computer vision tasks, deep learning has greatly
empowered semantic segmentation \cite{zhao2017pspnet,chen2017deeplab,long2015fully,noh2015learning,lin2017refinenet,chen2018encoder,zhen2018learning}
with its great representation learning ability.
These state-of-the-art methods mainly focus on 
the publicly available datasets like Pascal 
VOC \cite{everingham2010pascal},
ADE20K \cite{zhou2017scene}, CityScapes \cite{Cordts2016Cityscapes},
in which images are assumed to be 
independent and identically distributed.
However, this assumption does not stand in real-world scenarios.
For example, in mobile photography, a user may take
pictures to record his/her own life and form a personalized 
image set.
On the one hand, the personalized data does not have an identical 
distribution with public datasets, leading to a generalization issue 
when adopting well-trained segmentation models trained on public datasets.
On the other hand, as shown in \figref{figure:intro_fig},
images from the same user are correlated.
It yields potential studies on utilizing this
interrelated property to facilitate segmentation.

This paper addresses \textit{personalized image segmentation},
a problem that has not been discussed in previous works.
The difficulties mainly lie in the following two aspects:
(i) Firstly, 
there exists a large distribution gap 
between the public datasets and a user's personal data.
A simple method is to 
use extra annotations of the user's data to
train the model, 
which is very costly.
So it is urged to learn directly from unlabeled
personalized data.
However, there is no available personalized dataset
to learn.
(ii) Additionally,
the personalized images from the same user usually
have some personal traits. 
How to properly utilize these personalized traits in semantic
segmentation remains an unaddressed problem.
Despite the above difficulties, there are great demands for the 
personalized image segmentation in practice. 
For example, camera apps may need to generate high-quality segmentation 
masks for user's images to assist photography.
To tackle these challenges 
and open up further studies on personalized image 
segmentation,
we propose a personalized dataset called PSS and
a baseline personalized segmentation method based on
the proposed dataset.

The PSS dataset contains 15 individual users'
personalized images, resulting in 10080 images in total.
For each user's personalized data, we randomly select 
around 30\% of images
and annotate their pixel-level semantic masks for validation.
For easier adaptation from existing datasets to our
personalized data, 
we consider the 20 common object classes as in the PASCAL 
VOC\cite{everingham2010pascal} dataset.
To our knowledge, the PSS dataset is the first dataset
that focuses on the personalization issue of image
segmentation.
It enables researchers to study segmentation problems
by utilizing personalized traits.

On the challenge of learning with personal traits, 
previous studies on vision tasks' personalization 
issues
\cite{kim2020pienet, park2017attend, horiguchi2018personalized}
usually extract a global memory from personal data
to represent the user's preference or personality.
The extracted memory is then adopted as a prior for 
specific downstream tasks.
However, 
we consider personalization from a broader perspective,
\ie,
the personal traits lie in each person's
images, which is unnecessary to be extracted as a global
representation. 
In fact, experiments indicate that a global representation
extracted for a user fails in our scenario.
The failure is reasonable since in the semantic
segmentation problem,
we need to predict the class of each pixel for an image.
Though certain users' images have their characteristics,
there are still various classes of objects and scenes among
these images.
A global representation for all pixels would be too ambiguous.
To avoid the ambiguity of global representation while
learning to segment with personalized traits.
We propose to investigate the contextual connection between
personalized images and utilize them among correlated
images locally.
Specifically, 
we first cluster a person's images into several groups
so that images from the same group share similar
objects or backgrounds.
Then within each group, 
we extract multiple local region representations.
For the prediction of each pixel in that group,
we consult with correlated regional representations
using an attention mechanism.

Note that no labeled training images are provided in
the proposed personalized dataset.
We tackle this problem by domain adaptation from the existing
labeled dataset (as source) to the personalized images (as target).
There are many works on the problem of 
unsupervised domain adaption semantic segmentation (UDASS)
\cite{CLAN,BDL,SIBAN,LTIR,dcan,chang2019all,SSFDAN,lee2020unsupervised}.
While our personalized images are correlated
with each other,
current UDASS methods all treat target images as
independently distributed.
They can't capture the personality traits in the personal data.
In our baseline approach,
we incorporate a group context module into the
domain adaptation framework.
It allows the network to adapt from existing 
dataset to the personalized images while
taking advantage of the personal traits in
the personalized images.

The contributions of this work are two folds:
\begin{itemize}
	\item We first propose the personalization issue of image semantic
	segmentation and collect a personalized image dataset 
	called PSS, containing
	15 different
	users' data.
	\item We select some recent works related to this problem
	and report their performances
	on our dataset.
	Furthermore, we propose a baseline method
	that studies personal traits by
	learning local region representations.
	The approach achieves state-of-the-art performance on the 
	proposed personalized dataset.
\end{itemize}

\section{Related Work}
\subsection{Personalization Research}
Personalization issue has been discussed in many computer vision
and NLP tasks. \cite{mirkin2015motivating} use personality traits
to enhance the machine translation system. 
\cite{horiguchi2018personalized} proposed a personalized classifier
for food image classification.
\cite{park2017attend} predict caption and hashtag for social media
images by exploring personality from the user's previous posts.
\cite{kim2020pienet} studied image enhancement based on the user's
preference.
These methods usually focus on learning a global representation
from existing data, which serves as a prior when facing new data.
In this paper, we explore personalized image semantic
segmentation, a problem that has not been discussed previously.
In our proposed problem, personalization can be investigated
from the user's global trait,
and the personalized images' correlated property.

\subsection{Learning from Correlated Data}
A key challenge of the proposed personalized image segmentation
is to learn from the correlated images of the same person,
\ie, to extract complementary semantics while discarding
misleading semantics.
Co-segmentation~\cite{Hsu2019CVPR,Zhu2020CVPR,li2019group,han2017robust} 
and Co-saliency detection~\cite{fan2020taking,zhang2020gradient}
aim to mine the common semantic
objects in grouped images,
where each group contains the same class of objects.
Li~\etal~\cite{li2019group} proposed a recurrent network architecture
to explore the common semantic representation.
Zhang~\etal~\cite{zhang2020gradient} utilize the common classification features
to discover the general area of the target.
These methods usually learn
group representation for each group, which serves as
a prior when segmenting images from that group.
Our situation is more complex since the personalized data may contain
different objects in different images, and we need
to segment all these classes instead of just one.

\begin{figure*}[!ht]
	\centering
	\begin{overpic}[width=1\textwidth]{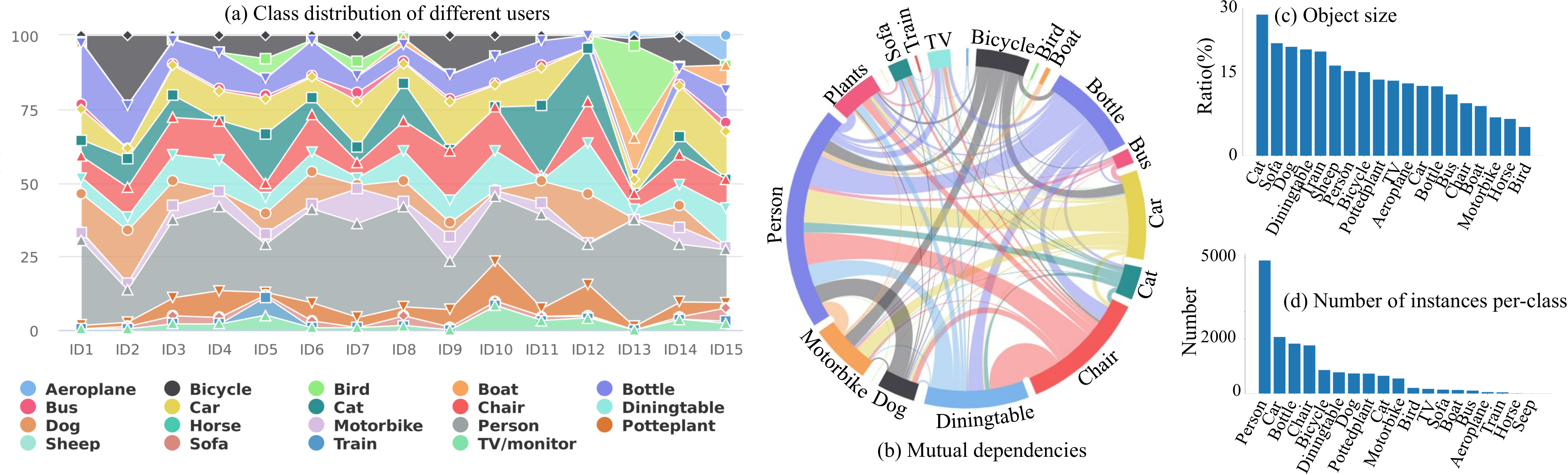}
	\end{overpic}
	\caption{Statistics of the PSS
		dataset. (a) The proportion of different 
		object classes of each user's 
		personalized data. (b) The mutual
		dependencies of different classes over the dataset. 
		(c) The average size of 
		different classes. (d) The number of 
		instances per-class.}
	\label{fig:fig}
\end{figure*}

\subsection{Domain Adaptation for Semantic 
	Segmentation}
The objective of personalized image segmentation is to
predict segmentation masks for unlabeled personalized images
with existing datasets and models.
A similar task that has been well studied recently is
unsupervised domain adaptation for semantic segmentation.
We will call it UDASS in the rest of the paper.
Given a labeled source dataset and unlabeled target dataset,
UDASS aims at tackling the distribution mismatch between the
source dataset and the target dataset and make the model generalize
well from source to target.
One line of work
\cite{vu2019advent,chen2019domain,tsai2019domain,tsai2018learning,hoffman2018cycada,pan2020unsupervised,pce,kim2020cross}
for UDASS use adversarial-based methods to align the distribution
shift of source and target domains.
Another line of work ~\cite{zhang2019curriculum,zhang2017curriculum,zou2018unsupervised,lian2019constructing,differential,PLCA,CCM,huang2020contextual} 
focus on learning strategy:
use curriculum learning or self-training strategies to babysit
the network towards learning good semantics for the target domain.
The main difference between our problem and UDASS is that we
do not consider images of the target domain independently.
Instead, we consider images from the same person as correlated parts.
Similar images may provide useful information for others.
Besides,
current UDASS methods 
mainly focus on urban scene datasets like
Cityscapes\cite{Cordts2016Cityscapes}, 
GTA5\cite{richter2016playing}, and SYNTHIA\cite{ros2016the},
where they do domain adaptation between synthetic images and
real images.
In the proposed dataset, we focus on personalized images of 
common objects.
Our dataset provides a more diverse and realistic personalized
scenario that can also evaluate the effectiveness of
domain adaptation approaches.



\subsection{Datasets for Semantic Segmentation}
As with other computer vision tasks, datasets play a key role
in the research of image segmentation.
Recent datasets 
have greatly empowered deep learning-based
segmentation frameworks
\cite{chen2017deeplab,zhao2017pspnet,he2017mask}.
PASCAL VOC \cite{everingham2010pascal} and
COCO \cite{lin2014microsoft} are datasets
focusing on images of common objects.
ADE20K \cite{zhou2017scene} also focuses on common objects
but with more fine-grained class labels like object parts.
CityScapes \cite{Cordts2016Cityscapes}, 
GTA5 \cite{richter2016playing},
SYNTHIA\cite{ros2016the},
and Synscapes\cite{wrenninge2018synscapes:}
are datasets of urban scenes.
Though many datasets have been proposed recently,
none of these datasets consider the personalization issue
in segmentation.
In this paper, we collect the PSS dataset from various users.
Our dataset concentrates on images of common objects with
different users' personalized traits.
It can be a good start for researches in personalized
image segmentation.
It can also provide a good benchmark for other segmentation
tasks like domain adaptation.

\section{Proposed Dataset}
\subsection{Dataset Collection}
To mimic real-world personalized data distribution, we directly
collect our dataset from different volunteers.
Each volunteer is asked to export images in
his/her mobile phone or camera to form his/her personal data.
For privacy, the volunteers are asked to look through the
images and filter out images that he/she is not willing to
make public.
Our dataset focuses on the 20 classes as in PASCAL VOC 
\cite{everingham2010pascal}.
Eventually, we get a large-scale dataset with 10080
images composed of 15 users'
personalized data.
Each personalized data can have different data distribution 
than
other users and may have some of its high-/low-level
statistics useful for semantic segmentation.

\subsection{Data Annotation}
We asked several well-trained experts to annotated the collected
personalized data.
Both image-level and pixel-level annotations
are provided in our dataset.

\noindent\textbf{Image-Level Annotation.} Consistency with
\cite{everingham2010pascal}, all images in our dataset are
labeled with the class label of objects that appeared.
On the one hand, the image-level labels can be used 
for data analysis.
On the other hand, the image-level labels may 
benefit the personalized semantic segmentation 
due to their success in many weakly supervised segmentation approaches 
\cite{liu2020leveraging,ahn2018learning,hou2018self,papandreou2015weakly,jiang2019integral}.
We show the object class distribution for each
user and the mutual dependencies of different classes in 
\figref{fig:fig}(a) and \figref{fig:fig}(b).
\vspace{5pt}

\noindent\textbf{Pixel-Level Annotation.} The challenge of
personalized image segmentation is to generate segmentation
masks on unlabeled personalized images. 
For model evaluation on our dataset,
we provide pixel-wise annotation for around 30\% of each
user's data.
For each pixel-wise annotated image, object regions belong to
the 20 classes.
As in PASCAL VOC~\cite{everingham2010pascal}, they are labeled with
certain values, resulting in a pixel-wise mask indicating the class
of every pixel in the image.
We show the average size and the number of instances for different classes
in \figref{fig:fig}(c) and \figref{fig:fig}(d).



\subsection{Dataset Characteristics}

\noindent\textbf{Personalized Data.} 
The most important 
characteristic of our dataset is personalization.
This naturally 
leads to intra-user coherency,
\ie, certain user's data has its
trait that might be coherent among different images,
which can be utilized to facilitate learning.
On the other hand,
different user's images differ in both low-level 
(\eg, light condition, picture quality) and
high-level
(\eg, image contents, background) properties.
The diversity of data distribution between different
users requires segmentation models to 
adjust to certain user's data.
More details of the intra-user coherency
and the inter-user distribution gap
can be found in the supplementary material.


\begin{figure*}[htbp]
	\centering
	\includegraphics[width=1.0\textwidth]{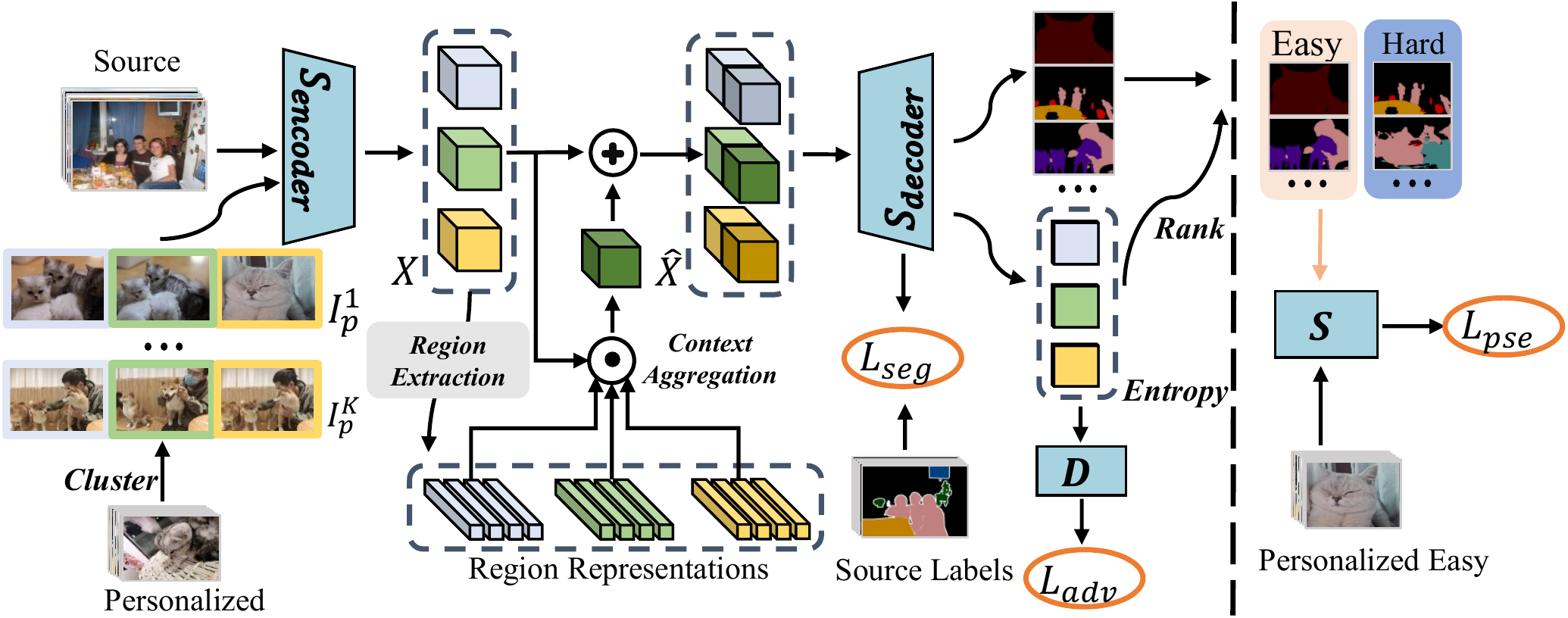}
	\caption{The pipeline of our approach for personalized
		image segmentation. Our model consists of two steps.
		The first step is the domain adaptation step, as in (a).
		In the second step, we further add a pseudo label
		loss $L_{pse}$ as in (b).
		In (a), we first cluster the personalized data into $K$ groups.
		Then in each group, we enhance the image representation
		$X$ with regional group context to obtain $\hat X$.
		For simplicity, we only show three images for each group,
		and we only show the group region context aggregation process for the 
		image tagged green.}
	\label{figure:pipline}\vspace{7pt}
\end{figure*}

\vspace{5pt}
\noindent\textbf{Realistic Data.}
Our personalized dataset is very close 
to realistic scenarios.
The realism lies in two folds.
%
%
Firstly, our dataset is directly collected from different users.
These images faithfully reflect what they care about and take pictures in daily life,
which means our dataset's results can reflect the effectiveness of different practice methods.
As the examples shown in the supplementary
material:
some users have more images of foods or
pets about their daily life,
while others have more images of beautiful scenery.
This indicates the importance of personalized segmentation.
%
Secondly, the object classes in our dataset are long-tailed
distributed, as illustrated in \figref{fig:fig}(d).
Some objects are more likely to be filmed
while others are not, \eg,
there might be \textit{"person"} in most of the images,
while only a few instances of \textit{"boat."}
How to settle the unbalanced class distribution problem
can be an interesting direction to explore.





\section{Proposed Approach}
In this section, we introduce our proposed baseline
method for personalized image segmentation.

\textbf{Overview.}
Consider source data with images 
$\{I_s \subset \mathbb{R}^{3\times H \times W}$ \}
and its C-class segmentation labels
$\{L_s \subset \mathbb{R}^{C\times H \times W}$\},
the unlabeled personalized data
$\{I_p \subset \mathbb{R}^{3\times H \times W}$\}.
Our approach's key idea is to utilize the correlation
between personalized images $\{I_p\}$
by using context from other images of the same user.
We show the architecture of our approach in \figref{figure:pipline}.
Our personalized image segmentation framework has two major steps:
a domain adaptation step and a pseudo label refinement step.
In the first step, 
we adapt from source data to personalized data with
an adversarial based domain adaptation framework.
During training, we incorporate our proposed group region
context module to utilize the inter-image context in 
personalized data.
In the second step, we select easy images
in the personalized data as pseudo labels with entropy maps.
The pseudo labels are used as ground-truth 
of the easy images to guide the segmentation network.

\subsection{Adversarial Based Domain Adaptation}
We start by introducing the adversarial based domain 
adaptation technique we used in step one.
Denoting a segmentation network as $S$, it takes image
$I_s$ as input and outputs a soft prediction map
$P_s = S(I_s) \in \mathbb{R}^{C \times H \times W}$,
where each value $P_{s}^{(c,h,w)}$ indicates the probability
that pixel $I_{s}^{(h,w)}$ belongs to class $c$.
Given $I_s$'s ground-truth $Y_s$, a cross-entropy 
loss:
\begin{equation}
    \mathcal{L}_{seg} = -\sum_{h,w} \sum_{c} Y_s^{c,h,w} log(P_{s}^{(c,h,w)})
\end{equation}

is optimized to train the segmentation network.
Besides the segmentation loss,
an adversarial training paradigm is adopted
to align the distribution discrepancy between 
source data $\{I_s\}$
and personalized data $\{I_p\}$.
Given the source and personalized image's segmentation
prediction $P_s$ and $P_p$, we compute their entropy
map by
\begin{equation}
    E_s^{h,w} = \sum_c -P_s^{c,h,w}log(P_s^{c,h,w}).
\end{equation}
A discriminator $D$ is trained to predict the domain
labels of $E_s$ and $E_p$.
By training the segmentation network $S$ to fool $D$,
we can close the distribution gap between the predictions
of source and personalized data.
The adversarial loss is formulated as:
\begin{equation}
    \mathcal{L}_{adv} (I_s, I_p) = -\sum_{h,w} log(1-D(E_s^{h,w})) + log(D(E_p^{h,w})).
\end{equation}

This adversarial paradigm can align the distribution
mismatch between source data and our personalized data.
However, it takes each image in personalized data individually,
thus fails to consider the correlation within $\{I_p\}$.
For this purpose,
we propose a group region context module to utilize the
inter-image context of the personalized data.

\subsection{Group Context Module}
We design a simple group context module to
utilize the correlated property of the proposed personalized
dataset.
We 
first cluster each user's personalized
data into multiple groups.
Each group contains images with similar semantics.
Within each group, we extract
soft region-wise context representations for all the images.
During segmentation, all soft region-wise context representations
are inferred to help training.
%

For a user's personalized data $\{I_p\}$,
we feed them into ResNet-50\cite{he2016deep} pretrained on
ImageNet\cite{deng2009imagenet} and get the representation
$\{F_p \in \mathbb{R}^{2048}\}$
before the last fully connected layer.
Then
we adopt the K-means clustering algorithm on
$\{F_p \}$,
obtaining $K$ groups of images as 
$\{\{I_p^{1}\}, \{I_p^{2}\}, \cdots, \{I_p^{K}\}\}$.
Consider the segmentation network as the composition of
encoder $S_{encoder}$ and decoder $S_{decoder}$.
The encoder takes image $I_{p}$ from group $k$ 
as input and outputs
intermediate representation
$X = S_{encoder}(I_{p}) \in \mathbb{R}^{CH \times W \times H}$,
$CH$ and $W, H$ indicate the channel and spatial size
of $X$, respectively.
Our group context module $F_{group}$
learns an enhanced representation
$\hat X = F_{group}(X) \in \mathbb{R}^{CH \times W \times H}$
by utilizing the group context.
There are two steps in the group context module:
\textit{region context extraction} and 
\textit{group region context aggregation}.

\textbf{Region context extraction.}
Inspired by \cite{yuan2019object}, we 
partition image $I_p$ into $C$ soft object regions.
$C$ is the number of object classes.
Using the aux output
$P_p \in \mathbb{R}^{C \times W \times H}$
of the segmentation network.
We compute each soft region's representation as
\begin{equation}
    f_c = \sum_{i} r_{ci} X_{i},
\end{equation}
where $i$ denotes the spatial location, $X_i$
represents pixel $i$.
$r_{ci}$ is the weight of pixel $i$ computed by
softmax normalization of $P_{pi}\in \mathbb{R}^{C}$ over the $C$ 
dimensions
as $r_{ci} = softmax(P_{pi})_c$.
For a group with $N$ images, we can extract 
$N\times C$ region representations for this group.

\definecolor{mygray}{gray}{0.88}
\begin{table*}[htbp]
\footnotesize
	\begin{center}
		\renewcommand\tabcolsep{3.4pt}
		\begin{tabular}{l|c|lllllllllllllll|l}
            \toprule[0.7pt]
            Method  & Backbone & \multicolumn{1}{c}{1} & \multicolumn{1}{c}{2} & \multicolumn{1}{c}{3} & \multicolumn{1}{c}{4} & \multicolumn{1}{c}{5} & \multicolumn{1}{c}{6} & \multicolumn{1}{c}{7} & \multicolumn{1}{c}{8} & \multicolumn{1}{c}{9} & \multicolumn{1}{c}{10} & \multicolumn{1}{c}{11} & \multicolumn{1}{c}{12} & \multicolumn{1}{c}{13} & \multicolumn{1}{c}{14} & \multicolumn{1}{c|}{15} & \multicolumn{1}{c}{Mean} \\ \hline
             No-DA & \multirow{7}{*}{ResNet-50} & 45.39 & 51.99 & 48.95 & 47.60 & 58.03 & 48.15 & 56.86 & 62.45 & 48.23 & 45.14 & 62.37 & 51.68 & 48.56 & 48.13 & 41.57 & 51.01 \\ 
            AdaptSeg~\cite{tsai2018learning} & & 46.87 & 52.16 & 50.06 & 48.51 & 59.78 & 51.39 & 57.12 & 63.41 & 50.99 & 46.15 & 60.68 & 52.84 & 50.32 & 50.69 & 43.08 & 52.27 \\ 
            MaxSquare~\cite{chen2019domain} & ~ & 48.28 & 52.50 & 50.61 & 50.54 & 61.39 & 54.60 & 59.36 & 63.43 &
            50.67 & 46.49 & 62.94 & 52.68 & 49.65 & 48.99 & 46.00 & 53.20\\
            FDA~\cite{yang2020fda} & & 50.12 & 53.70 & 53.22 & 50.76 & 60.29 & 55.01 & 58.18 & 65.89 & 53.28 & 46.49 & 62.09 & 56.10 & 48.93 & 51.38 & 47.03 & 54.16\\ 
            ADVENT~\cite{vu2019advent} & & 53.39 & 57.33 & 52.42 & 52.51 & 64.63 & 55.04 & 60.61 & 61.69 & 55.34 & 49.18 & 66.05 & 57.83 & 56.04 & 54.38 & 52.34 & 56.59 \\ 
            MRNet~\cite{zheng2020unsupervised} & & \textbf{54.05} & 58.62 & 54.29 & 53.17 & 61.72 & 57.24 & 62.20 & 66.46 & 56.75 & 50.27 & 66.76 & 54.20 & 53.87 & 54.38 & 51.38 & 57.02 \\ \cline{1-1} \cline{3-18}
            OURS-S1 & & 52.90 & 59.12 & 54.74 & 55.82 & 64.97 & \textbf{60.38} & 61.78 & 68.12 & 56.99 & 51.21 & 69.42 & 60.44 & 57.05 & 54.41 & 54.51 & 58.79 \\
            OURS-S2 & & 53.28 & \textbf{60.39} & \textbf{54.81} & \textbf{56.02} & \textbf{66.87} & 60.11 & \textbf{63.77} & \textbf{69.09} & \textbf{57.44} & \textbf{52.66} & \textbf{70.42} & \textbf{60.77} & \textbf{58.50} & \textbf{56.84} & \textbf{54.85} & \textbf{59.72} \\
            \hline 
            No-DA & \multirow{7}{*}{VGG-16} & 33.68 & 33.56 & 35.50 & 35.49 & 39.52 & 37.55 & 36.23 & 47.95 & 34.35 & 32.86 & 50.95 & 41.48 & 39.24 & 30.90 & 34.51 & 37.58 \\
            AdaptSeg~\cite{tsai2018learning} & & 32.70 & 37.65 & 37.16 & 33.54 & 40.55 & 41.11 & 43.17 & 52.12 & 36.95 & 31.83 & 49.04 & 40.97 & 33.54 & 31.49 & 34.06 & 38.39 \\
             MaxSquare~\cite{chen2019domain} & ~ & 36.17 & 32.99 & 38.81 & 37.36 & 42.64 & 42.03 & 49.88 & 50.06 &
            37.99 & 35.93 & 51.33 & 41.98 & 36.27 & 36.35 & 37.13 & 40.46 \\
            FDA~\cite{yang2020fda} & & 34.61 & 36.75 & 35.53 & 36.60 & 38.36 & 40.07 & 45.21 & 52.57 & 37.79 & 35.01 & 49.59 & 41.93 & 33.72 & 35.01 & 36.27 & 39.27 \\
            ADVENT~\cite{vu2019advent} & & 39.89 & 44.39 & 39.88 & 40.01 & 49.89 & 44.24 & 47.99 & 54.59 & 43.84 & 38.29 & 53.00 & 43.07 & 42.83 & 40.02 & 41.36 & 44.22 \\
            MRNet~\cite{zheng2020unsupervised} & & 34.40 & 41.18 & 36.67 & 32.18 & 44.63 & 38.12 & 41.99 & 46.78 & 39.51 & 36.54 & 39.39 & 44.17 & 35.93 & 37.17 & 38.35 & 39.13 \\
            \cline{1-1} \cline{3-18}
             OURS-S1 & & 41.87 & 45.73 & 43.14 & \textbf{44.04} & 52.44 & 47.45 & \textbf{52.32} & 56.92 & 45.61 & \textbf{42.67} & 54.94 & \textbf{48.38} & 44.24 & 41.67 & 45.98 & 47.16 \\
            OURS-S2 & & \textbf{43.24} & \textbf{47.89} & \textbf{44.67} & 44.00 & \textbf{53.27} & \textbf{50.68} & 52.18 & \textbf{57.86} & \textbf{46.84} & 42.34 & \textbf{56.56} & 46.28 & \textbf{47.02} & \textbf{42.98} & \textbf{47.01} & \textbf{48.19} \\
            \toprule[0.7pt]
        \end{tabular}
	\end{center}
	\vspace{-5pt}
	\caption{FIoU results for different
	methods with ResNet-50 \cite{he2016deep} and VGG-16 \cite{simonyan2014very}.
	The column number indicates the 15 user IDs. The column ”Mean” denotes the mean performance overall IDs.
	Best results are highlighted in \textbf{bold}.}
	\label{table:benchmark__fiou}
\end{table*}

\definecolor{mygray}{gray}{0.88}
\begin{table*}[htbp]
\footnotesize
	\begin{center}
		\renewcommand\tabcolsep{3.4pt}
		\begin{tabular}{l|c|lllllllllllllll|l}
			\toprule[0.7pt]
			Method  & Backbone & \multicolumn{1}{c}{1} & \multicolumn{1}{c}{2} & \multicolumn{1}{c}{3} & \multicolumn{1}{c}{4} & \multicolumn{1}{c}{5} & \multicolumn{1}{c}{6} & \multicolumn{1}{c}{7} & \multicolumn{1}{c}{8} & \multicolumn{1}{c}{9} & \multicolumn{1}{c}{10} & \multicolumn{1}{c}{11} & \multicolumn{1}{c}{12} & \multicolumn{1}{c}{13} & \multicolumn{1}{c}{14} & \multicolumn{1}{c|}{15} & \multicolumn{1}{c}{Mean} \\ \hline
			No-DA & \multirow{7}{*}{ResNet-50} & 28.05 & 29.18 & 30.78 & 33.05 & 42.52 & 31.31 & 35.85 & 28.63 & 39.60 & 36.99 & 33.15 & 38.51 & 29.78 & 32.75 & 31.85 & 33.47 \\ 
			AdaptSeg~\cite{tsai2018learning} & & 31.69 & 28.87 & 30.50 & 35.09 & 45.83 & 32.55 & 36.70 & 33.83 & 36.43 & 36.49 & 34.09 & 41.23 & 31.02 & 35.52 & 34.40 & 34.95 \\ 
			MaxSquare~\cite{chen2019domain} & & 28.72 & 28.91 & 31.81 & 36.45 & 40.09 & 33.94 & 38.85 & 31.21 & 35.85 & 32.23 & 28.58 & 34.16 & 33.58 & 30.35 & 34.78 & 33.30 \\ 
			FDA~\cite{yang2020fda} & & 31.94 & 31.16 & 32.39 & 36.11 & 45.35 & 35.76 & 37.46 & 30.93 & 42.91 & \textbf{43.28} & \textbf{37.51} & 38.09 & 29.31 & \textbf{37.25} & 35.76 & 36.35\\ 
			ADVENT~\cite{vu2019advent} & & 36.04 & 34.04 & 36.98 & 39.98 & 43.76 & 40.52 & 41.59 & 29.69 & 36.26 & 39.19 & 33.46 & 39.05 & 38.17 & 33.43 & 37.44 & 37.31 \\ 
			MRNet~\cite{zheng2020unsupervised} & & \textbf{38.27} & \textbf{35.02} & 36.98 & 36.54 & 43.99 & \textbf{40.90} & 40.22 & 36.26 & 32.35 & 33.10 & 36.26 & 31.78 & 37.77 & 35.89 & 32.24 & 36.51 \\ 
			\cline{1-1} \cline{3-18}
			 OURS-S1 & & 36.61 & 34.43 & 31.88 & 40.36 & 44.25 & 33.64 & 38.14 & 32.25 & 39.87 & 38.69 & 37.20 & 42.44 & \textbf{39.60} & 30.37 & \textbf{42.18} & 37.46 \\
            OURS-S2 & & 33.85 & 33.38 & \textbf{38.40} & \textbf{41.36} & \textbf{46.73} & 37.58 & \textbf{44.19} & \textbf{36.87} & \textbf{44.66} & 42.03 & 37.42 & \textbf{43.71} & 35.12 & 34.18 & 37.89 & \textbf{39.16} \\
			\hline
			 No-DA & \multirow{7}{*}{VGG-16} & 15.78 & 17.19 & 17.80 & 21.41 & 21.54 & 18.35 & 19.07 & 15.40 & 21.67 & 22.80 & 18.55 & 21.06 & 21.66 & 18.96 & 22.95 & 19.61 \\
			AdaptSeg~\cite{tsai2018learning} & & 16.59 & 19.04 & 18.96 & 23.81 & 21.43 & 23.12 & 25.47 & 16.26 & 23.33 & 22.60 & 19.08 & 20.20 & 22.12 & 20.10 & 24.30 & 21.09 \\
			 MaxSquare~\cite{chen2019domain} & & 18.46 & 18.19 & 18.46 & 22.29 & 26.01 & 23.88 & 25.36 & 17.07 & 25.01 & 25.37 & 19.99 & 20.56 & 24.52 & 20.82 & 25.90 & 22.13 \\ 
			FDA~\cite{yang2020fda} & & 17.17 & 17.82 & 20.07 & 24.44 & 23.82 & 25.69 & 25.22 & 15.56 & 23.31 & 25.53 & 21.14 & 20.68 & 20.82 & 19.26 & 24.65 & 21.68 \\
			ADVENT~\cite{vu2019advent} & & 27.32 & 21.95 & \textbf{25.21} & 25.46 & 35.50 & 23.75 & 28.18 & \textbf{25.03} & \textbf{32.47} & 29.73 & 24.76 & 25.00 & 26.74 & 24.76 & 26.31 & 26.81 \\
			MRNet~\cite{zheng2020unsupervised} & & 20.30 & 16.46 & 20.92 & 27.62 & 29.70 & 25.46 & 27.74 & 22.30 & 30.64 & 26.46 & 17.64 & 27.12 & 23.43 & 23.86 & 26.08 & 24.38 \\
			\cline{1-1} \cline{3-18}
			 OURS-S1 & & 24.40 & \textbf{24.93} & 20.31 & 31.01 & 35.54 & 30.67 & 28.81 & 20.68 & 27.45 & \textbf{32.06} & 27.63 & \textbf{31.44} & 27.68 & 23.87 & \textbf{33.53} & 28.00 \\
            OURS-S2 & & \textbf{25.53} & 24.40 & 22.62 & \textbf{33.55} & \textbf{35.73} & \textbf{31.86} & \textbf{32.14} & 21.84 & 28.62 & 30.57 & \textbf{30.26} & 24.97 & \textbf{28.82} & \textbf{25.25} & 31.91 & \textbf{28.54} \\
			\toprule[0.7pt]
		\end{tabular}
	\end{center}
	\vspace{-5pt}
	\caption{MIoU results for different
	methods with ResNet-50 \cite{he2016deep} and VGG-16 \cite{simonyan2014very}.
	The column number indicates the 15 user IDs. The column ”Mean” denotes the mean performance overall IDs. 
	Best results are highlighted in \textbf{bold}. }
	\label{table:benchmark_miou}
\end{table*}

\definecolor{mygray}{gray}{0.3}
\begin{table*}[htbp]
\footnotesize
	\begin{center}
		\renewcommand\tabcolsep{5pt}
		\begin{tabular}{c|c c c c c c c c c c c c c c c| c}
			\toprule[0.5pt]
			Methods & 1 & 2 & 3 & 4 & 5 & 6 & 7 & 8 & 9 & 10 & 11 & 12 & 13 & 14 & 15 & Mean \\
			\hline
			None          & 39.89 & 44.39 & 39.88 & 40.01 & 49.89 & 44.24 & 47.99 & 54.59 & 43.84 & 38.29 & 53.00 & 43.07 & 42.83 & 40.02 & 41.36 & 44.22 \\
			Global        & 38.89 & 42.13 & 42.13 & 38.96 & 50.88 & 45.09 & 51.62 & 55.67 & 44.23 & 38.51 & 49.80 & 45.70 & 43.20 & 39.84 & 41.71 & 44.56 \\
			\hline
			OURS          & 41.87 & 45.73 & 43.14 & 44.04 & 52.44 & 47.45 & 52.32 & 56.92 & 45.61 & 42.67 & 54.94 & 48.38 & 44.24 & 41.67 & 45.98 & 47.16 \\
			\toprule[0.5pt]
		\end{tabular}
	\end{center}
	\vspace{-5pt}
	\caption{Ablation of the group context module.
	    "None" and "Global" denote no context and
	    global context, respectively.
        }
	\label{table:ab_ctx}
\end{table*}

\definecolor{mygray}{gray}{0.3}
\begin{table*}[htbp]
\footnotesize
	\begin{center}
		\renewcommand\tabcolsep{5pt}
		\begin{tabular}{c|c c c c c c c c c c c c c c c| c}
			\toprule[0.5pt]
			Groups & 1 & 2 & 3 & 4 & 5 & 6 & 7 & 8 & 9 & 10 & 11 & 12 & 13 & 14 & 15 & Mean \\
			\hline
			1     & 42.39 & 46.11 & 42.95 & 43.79 & 52.45 & 46.15 & 51.23 & 55.88 & 45.57 & 42.51 & 55.30 & 47.89 & 43.66 & 39.54 & 44.12 & 46.64 \\
			10    & 42.49 & 45.52 & 42.75 & 44.07 & 52.64 & 46.54 & 50.85 & 56.52 & 45.81 & 42.25 & 54.75 & 46.86 & 44.62 & 41.18 & 45.36 & 46.81 \\
			80    & 41.87 & 45.73 & 43.14 & 44.04 & 52.44 & 47.45 & 52.32 & 56.92 & 45.61 & 42.67 & 54.94 & 48.38 & 44.24 & 41.67 & 45.98 & 47.16 \\
			200   & 42.11 & 45.66 & 43.14 & 43.33 & 52.08 & 45.85 & 52.05 & 56.51 & 45.78 & 42.13 & 53.75 & 46.73 & 43.80 & 41.59 & 45.35 & 46.66 \\
			\toprule[0.5pt]
		\end{tabular}
	\end{center}
	\vspace{-5pt}
	\caption{Ablation study of different numbers of groups.
		Columns indicate
		different user IDs. FIoU is reported.}
	\label{table:ab_ngp}
\end{table*}

\definecolor{mygray}{gray}{0.3}
\begin{table*}[ht!]
\footnotesize
	\begin{center}
		\renewcommand\tabcolsep{5pt}
		\begin{tabular}{c|c c c c c c c c c c c c c c c| c}
			\toprule[0.5pt]
			Methods & 1 & 2 & 3 & 4 & 5 & 6 & 7 & 8 & 9 & 10 & 11 & 12 & 13 & 14 & 15 & Mean \\
			\hline
			MixSample     & 40.92 & 43.18 & 40.73 & 40.55 & 49.84 & 46.18 & 50.42 & 57.22 & 42.92 & 39.39 & 54.26 & 45.83 & 43.67 & 38.74 &  44.90 & 45.25 \\
			MixAll        & 42.54 & 45.11 & 41.73 & 39.87 & 49.58 & 43.74 & 52.64 & 56.88 & 43.82 & 38.55 & 55.26 & 47.18 & 42.33 & 38.97 &  45.16 & 45.56 \\
			\hline
			Personal       & 41.87 & 45.73 & 43.14 & 44.04 & 52.44 & 47.45 & 52.32 & 56.92 & 45.61 & 42.67 & 54.94 & 48.38 & 44.24 & 41.67 &  45.98 & 47.16 \\
			\toprule[0.5pt]
		\end{tabular}
	\end{center}
	\vspace{-3pt}
	\caption{Experiments on the mixed image set. "MixAll" denotes
	mix all the user's image, "MixSample" samples 1/15 from "MixAll" to have a similar size with each personalized data.}
	\label{table:ab_mix}
\end{table*}

\textbf{Group region context aggregation.} 
Given a group's region representations
$\{f_{i,j} | i\in [1,C], j \in [1,N]\}$,
we compute group context representation for each
pixel in $X$ by weighted aggregation of group regions:
\begin{equation}
    c_{h,w} = \rho(\sum_{i,j} w_{(i,j),(h,w)} \sigma(f_{i,j})).
\end{equation}
Here, $\rho$ and $\sigma$ are two linear transformation functions.
The weight $w_{(\tilde i,\tilde j),(h,w)}$ is computed by measuring
the relation between the pixel $X_{h,w}$ and region 
representation $f_{\tilde i,\tilde j}$ as
\begin{equation}
    w_{(\tilde i,\tilde j),(h,w)} = \frac{e^{s(X_{h,w}, f_{\tilde i,\tilde j})}}{\sum_{i\in [1,K], j \in [1,N]} e^{s(X_{h,w}, f_{i,j})}},
\end{equation}
where $s(X_{h,w}, f_{i,j})$ is a relation function formulated as
$s(X_{h,w}, f_{i,j}) = \phi(X_{h,w})^T \varphi(f_{i,j})$,
$\phi$ and $\varphi$ are two transform functions
implemented with one fully connected layer.

After obtaining the group context, we can enhance the
pixel representation as:
\begin{equation}
    \hat X_{h,w} = \psi ([X_{h,w}, c_{h,w}]).
\end{equation}
$[*,*]$ denotes concatenation, and $\psi$ is a linear transformation.
The result representation $\hat X$ will be fed
into the decoder and output the prediction map:
${\hat P_p} = S_{decoder}(\hat X)$.
For each pixel in $X$,
the group region context enhancing module aggregates representations
of similar regions in the same group as the group context,
which provides extra information for the segmentation network.

\subsection{Refine with Pseudo Label}
Besides a first step domain adaptation,
recent domain adaptation methods
\cite{pan2020unsupervised,zheng2019unsupervised}
for semantic segmentation
usually adopt pseudo labels to refine
the network further.
We also adopt this training paradigm with the
predictions of personalized images in our approach.
The entropy map $E_p$ introduced previously
is an indicator of the uncertainty of the segmentation
network for image $I_p$.
Prediction with low uncertainty usually means
the input image is simple, and the result has
high reliability.
So we choose predictions with low entropy
values as pseudo labels.
Note that each person's personalized data is
relatively small compared with datasets like 
VOC and CityScapes, which alone
doesn't have enough data for training the segmentation
network. 
So unlike in \cite{pan2020unsupervised}, we add the
pseudo labels to the network with an extra segmentation
loss $\mathcal{L}_{pse}$ instead of replacing the
source dataset with pseudo labels.


\section{Experiments}
\subsection{Datasets and Evaluation Metrics}
We collect our personalized dataset to have the 
same classes as PASCAL VOC\cite{everingham2010pascal}.
So during training, we use the augmented VOC training
set as the source dataset, which has 10582 labeled images
with 20 classes of objects.
Mean Intersection over Union (MIoU) is adopted for quantitative
evaluation.
Note that the personalized data are usually long-tailed distributed,
which means the classes are very unbalanced.
MIoU may be distorted due to this unbalance.
So we further use another metric called
Foreground Intersection over Union (FIoU).
FIoU reflects the mean IoU over images instead of classes.
Specifically, we first compute the foreground IoU
$IoU_i$ for
image $i$, then compute the mean IoU of all images
$\sum_{i}^{N}{IoU_i}$.

\subsection{Implementation Details}
We use ResNet50 \cite{he2016deep} pretrained from
ImageNet \cite{deng2009imagenet} as the backbone of the
segmentation network.
A PSP module \cite{zhao2017pspnet} is equipped to
the segmentation network as in \cite{vu2019advent}.
The inputs for adaptation training are source images
and labels, grouped target images.
To simplify training and save computation, we
do not use all image regions in a group to build
the group context. Instead, we constrain each batch
to be in the same group, then use the images
in each batch to compute it.
The random crop is adopted for image augmentation.
All the inputs are resized to $320\times 320$ during
training.
In the pseudo label refinement step, a select rate
of $r=0.5$ is used for choosing reliable predictions.
The masked pseudo label pixels are set to 255.
To simplify the training process and save GPU memory,
we do not process a whole group at one iteration.
Instead, we just make sure all images of a bath
are from the same group.
The batch size of all experiments is set to 8 in this paper.
We use the SGD optimizer \cite{bottou2010large} with
a learning rate of $2.5\times 10^{-4}$, 
The momentum and weight decay 
are set to 0.9 and $10^{-4}$, respectively.
The codes are implemented with PyTorch \cite{NEURIPS2019_9015}
library.

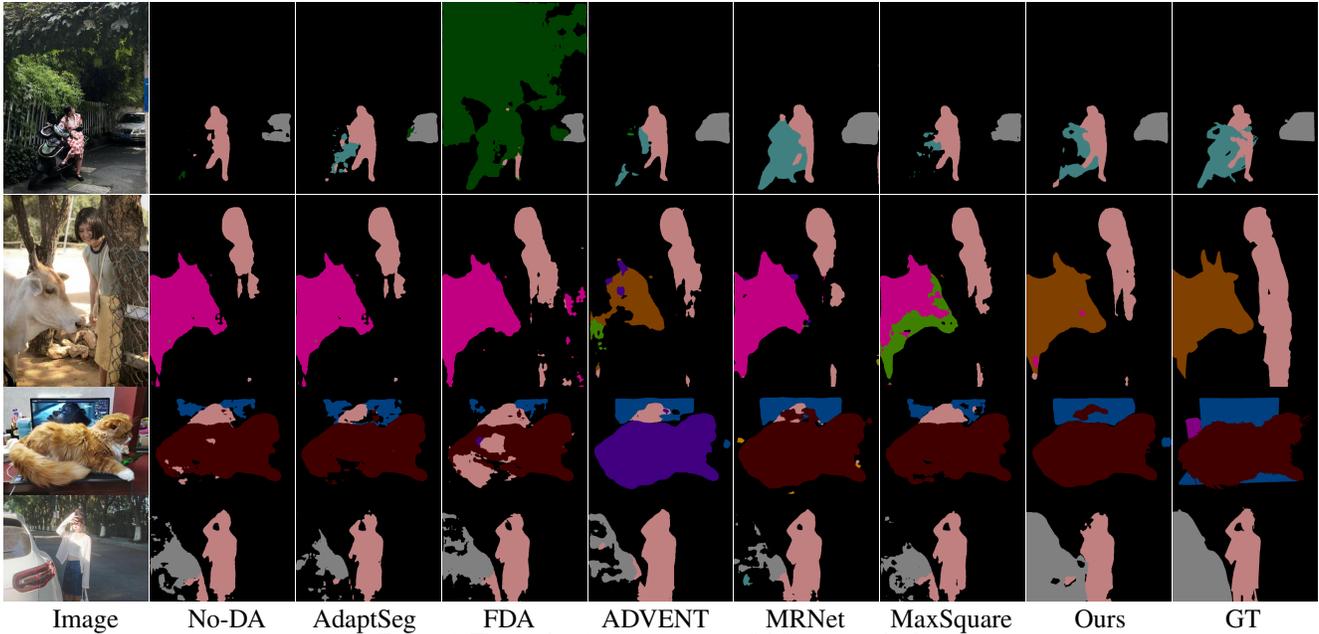
\begin{figure*}[htbp]
	\centering
	\begin{overpic}[width=\textwidth]{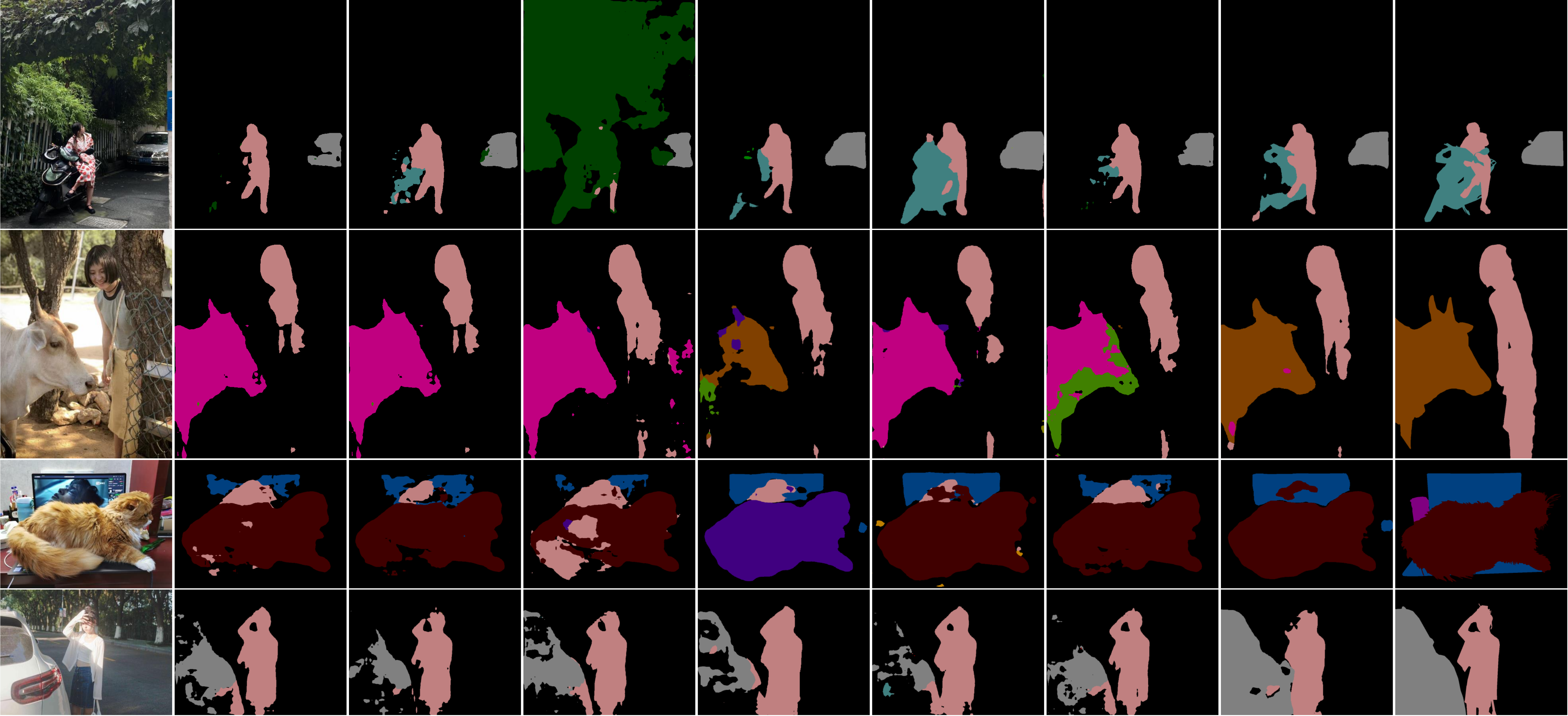}
		\put(3.7, -2){Image}
		\put(14, -2){No-DA}
		\put(23.5, -2){AdaptSeg}
		\put(36.5, -2){FDA}
		\put(45.5, -2){ADVENT}
		\put(58,  -2){MRNet}
		\put(67.5, -2){MaxSquare}
		\put(81.5, -2){Ours}
		\put(93, -2){GT}
	\end{overpic} \\
	\vspace{7pt}
	\caption{The qualitative comparison of different methods. }
	\label{figure:result}
\end{figure*}

\subsection{Performace Comparison}
We report the performances of some selected domain
adaptation methods on our dataset,
including 
AdaptSeg\cite{tsai2018learning},
MaxSquare\cite{chen2019domain},
FDA\cite{yang2020fda},
ADVENT\cite{vu2019advent}, and
MRNet\cite{zheng2020unsupervised}.
All these methods treat target images individually
without considering the correlated property of personalized
images.
All models are trained with
VOC\cite{everingham2010pascal} as the 
source and personalized data as the target.
Methods like MRNet\cite{zheng2020unsupervised} only
use the target pseudo labels to supervise the
segmentation network in step 2, which leads to poor performance
since the number of our personalized data is relatively small.
So we add extra supervision of VOC\cite{everingham2010pascal}
label for such methods.
The results are tested on the annotated validation split
of the personalized dataset.
We report the results of FIoU in
\tabref{table:benchmark__fiou}
and MIoU in \tabref{table:benchmark_miou}.
We denote our method without pseudo label refinement
and our full model as \textit{OURS-S1} and
\textit{OURS-S2}, respectively.

Overall, with ResNet50\cite{he2016deep} as the backbone,
\textit{OURS-S1} obtains 37.46 MIoU and 58.79
FIoU. Compared to the baseline method
ADVENT, it improves the performance by 0.15
and 2.20, respectively, which indicates the
effectiveness of our group context module.
Note that the MIoU improvement of 0.15
is relatively slight compared with FIoU.
We conjecture it is caused by the long-tailed
property of the personalized data.
Since the group context module incorporates
other images' context to help to learn, it
tends to perform better on classes with many 
images while may damage the results for rare
classes.
When evaluating MIoU, the results
can be affected by these rare classes.
We provide the class IoU results for different
users in the supplementary material.
By utilizing pseudo labels, \textit{OURS-S2}
obtains 39.16 MIoU and 59.72 FIoU, 
further improves
the performance by 1.7 and 0.93, respectively.
We show some of the predicted masks in \figref{figure:result}.

\section{Discussion}

\subsection{Effectiveness of Group Context}
In this section, we study the effectiveness
of our proposed group context module by
comparing it with two baselines:
\textit{None} and \textit{Global}.
\textit{None} refers to directly use the
feature $X$ from the encoder without context.
\textit{Global} denotes 
using a global group context
to enhance representation as in object
co-segmentation methods \cite{li2019group}.
Experiments here are conducted with 
backbone VGG-16
\cite{simonyan2014very}.
As reported in \tabref{table:ab_ctx},
the \textit{None} baseline achieves 44.22 FIoU
on average.
\textit{Global} slightly improves the performance
by 0.34, which indicates that a global group representation
is not effective enough in our situation.
\textit{OURS} improves the performance by 2.94,
which shows the effectiveness of the proposed
group context module.

\subsection{Significance of Personalized Training}
In this section,
we merge all the images from all the users to form
a large image set \textit{MixAll}, a subset of 672 images
\textit{MixSample}
is then randomly sampled from \textit{MixAll}.
We train our model on these image sets and 
evaluate the model on different users' data.
Results are reported in \tabref{table:ab_mix}.
With around 15 times of target images, the 
\textit{MixAll} achieves 45.56 FIoU, which
is lower than 47.16 of \textit{Personal}, \ie, 
training from corresponding personalized data.
The result shows the value of learning
from personalized data.


\subsection{Number of Groups}
In this section, we cluster each user's personalized
data into different numbers of groups and
investigate how the number of groups influences
the segmentation performance at inference time.
As in \tabref{table:ab_ngp}.
Different rows indicate different numbers of groups.
With \textit{Groups=1}, all images of certain users
are treated as in one group. When computing a group
region context, irrelevant images might be considered
and confusing the network.
With \textit{Groups=200}, the number of images in
each group is too small, thus can't provide enough
context for the group context module.
On average, we get a better FIoU of 47.16
with a group number of 80 compared with other
numbers.
However, we may still notice that different
users have the best result with different group numbers.
We conjecture that it is caused by the distribution
gap between different users.
%
The results indicate that we need different numbers
of groups for different users.
In the future, we'll study more flexible methods to
cluster the personalized images rather than using a
fixed group number.


\section{Conclusion}
In this paper, we address the personalization issue
in image semantic segmentation.
We first collect a large personalized image dataset
PSS with 15 users' data.
Our dataset can be a good start for investigating
the personalization issue in segmentation.
The challenges of the personalized image segmentation
problem are two folds.
One is how to learn from different users' unlabeled data;
another is how to utilize the personalized traits in
certain user's data.
By utilizing the personalized images' correlated property,
we propose a baseline method that
adopts the inter-image context to facilitate segmentation.
For future work, we will explore more sophisticated ways
to learn from the unlabeled data.
We will also investigate how to improve the performance of the
group context module in rare classes.

\paragraph{Acknowledgement.}

This research was supported by NSFC (61922046), 
S\&T innovation project from Chinese Ministry of Education,
BNRist(No.BNR2020KF01001)
and the Fundamental Research Funds for the Central Universities 
(Nankai University, NO. 63213090).

{\small
\bibliographystyle{ieee_fullname}
\bibliography{egbib}
}

\input{appendix}

\end{document}

%% file: appendix.tex
\clearpage
\appendix
\section{Appendix}
\subsection{More Dataset Information}
In Section 3.3 of the main paper,
we talk about
the intra-user coherency and inter-user difference
of the proposed dataset. 
In this section, we show more information of the proposed 
dataset for readers
to have more illustrative understanding of the personalized 
dataset.

\begin{figure}[th!]
	\centering
	\subfloat[as]{
		\begin{overpic}[width=.225\textwidth]{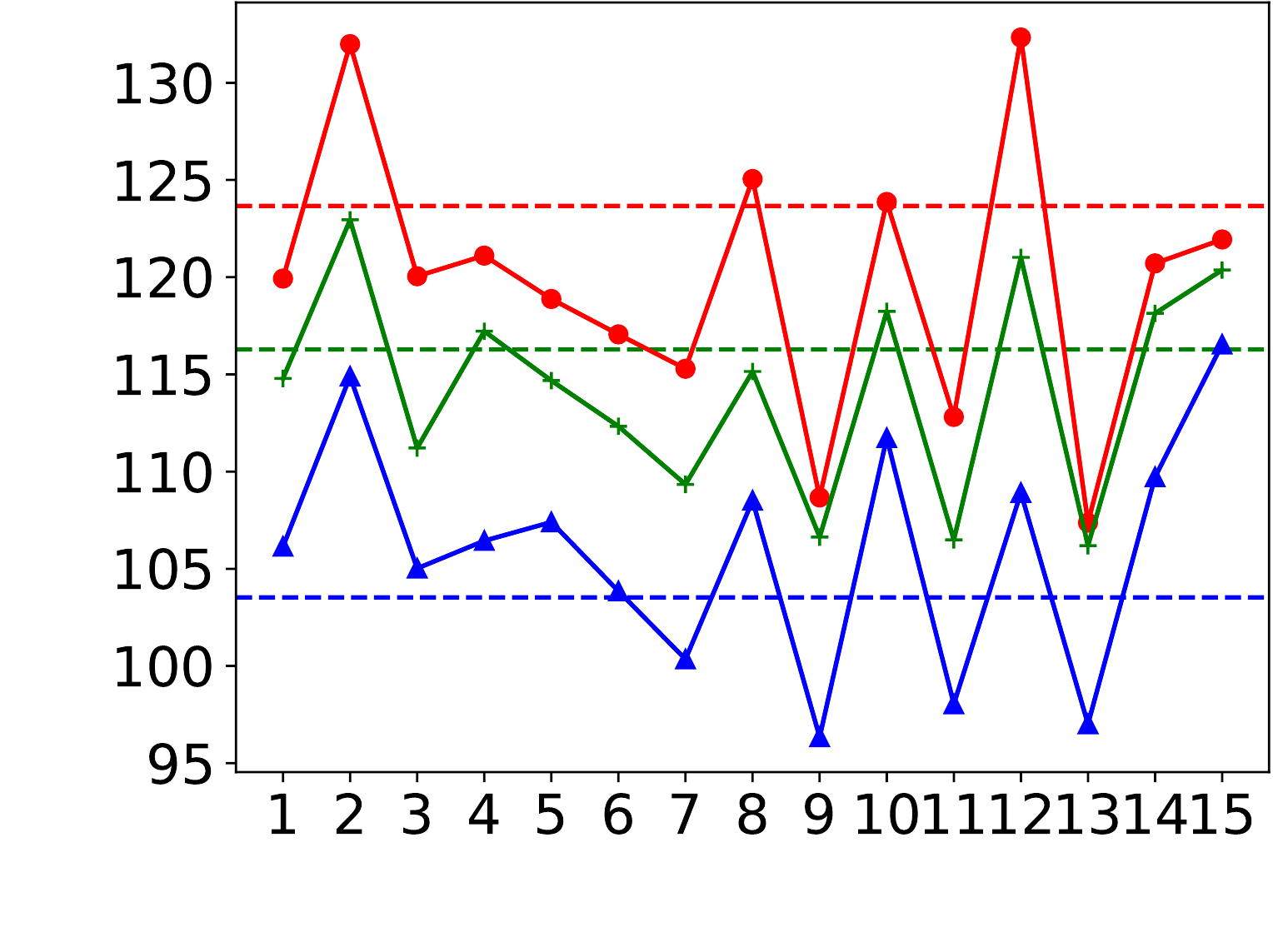}
			\put(0, 26){\rotatebox{90}{Intensity}}
			\put(50, 0){ID}
		\end{overpic}
	}
	\subfloat[bs]{
		\begin{overpic}[width=.225\textwidth]{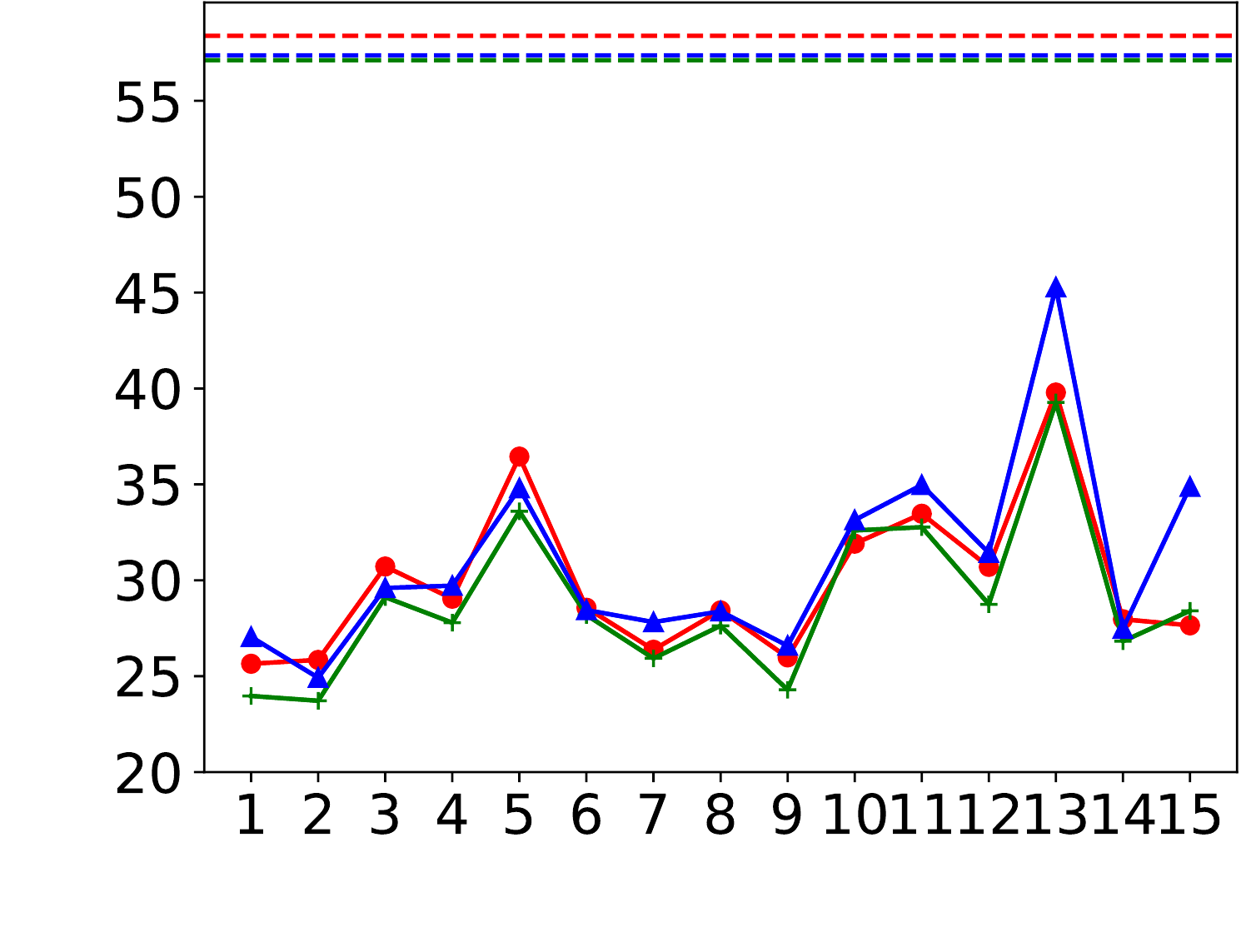}
			\put(50, 0){ID}
		\end{overpic}
	}
	\vskip -0.1in
	\caption{Mean Intensity (a) and its standard deviation (b) of our proposed dataset among different users.
	We use \textcolor{red}{red}, \textcolor{green}{green}, and \textcolor{blue}{blue} to represent different channels.
	The dotted lines denote the mean intensity and its standard deviation of ImageNet~\cite{imagenet}.}     \label{fig:meanandstd}
\end{figure}

\textbf{Low-level statistic.}
To show the low-level statistics of the dataset,
we calculate the mean and standard deviation of images 
of each user's images  in~\figref{fig:meanandstd}.
From both the mean and the standard deviation figure,
we can observe obvious differences between different users.
It reflects the inter-user difference between different
users in low-level statistics.
In \figref{fig:meanandstd}(b), we can also notice that
all users' images have
obviously lower deviation than ImageNet\cite{imagenet} 
dataset.
It indicates that 
images from personalized data tend to be more correlated with 
each other compared with
public datasets.
It also reflects the intra-user coherency of personalized
images in low-level statistics.

\begin{figure*}[t]
	\centering
	\includegraphics[width=\textwidth]{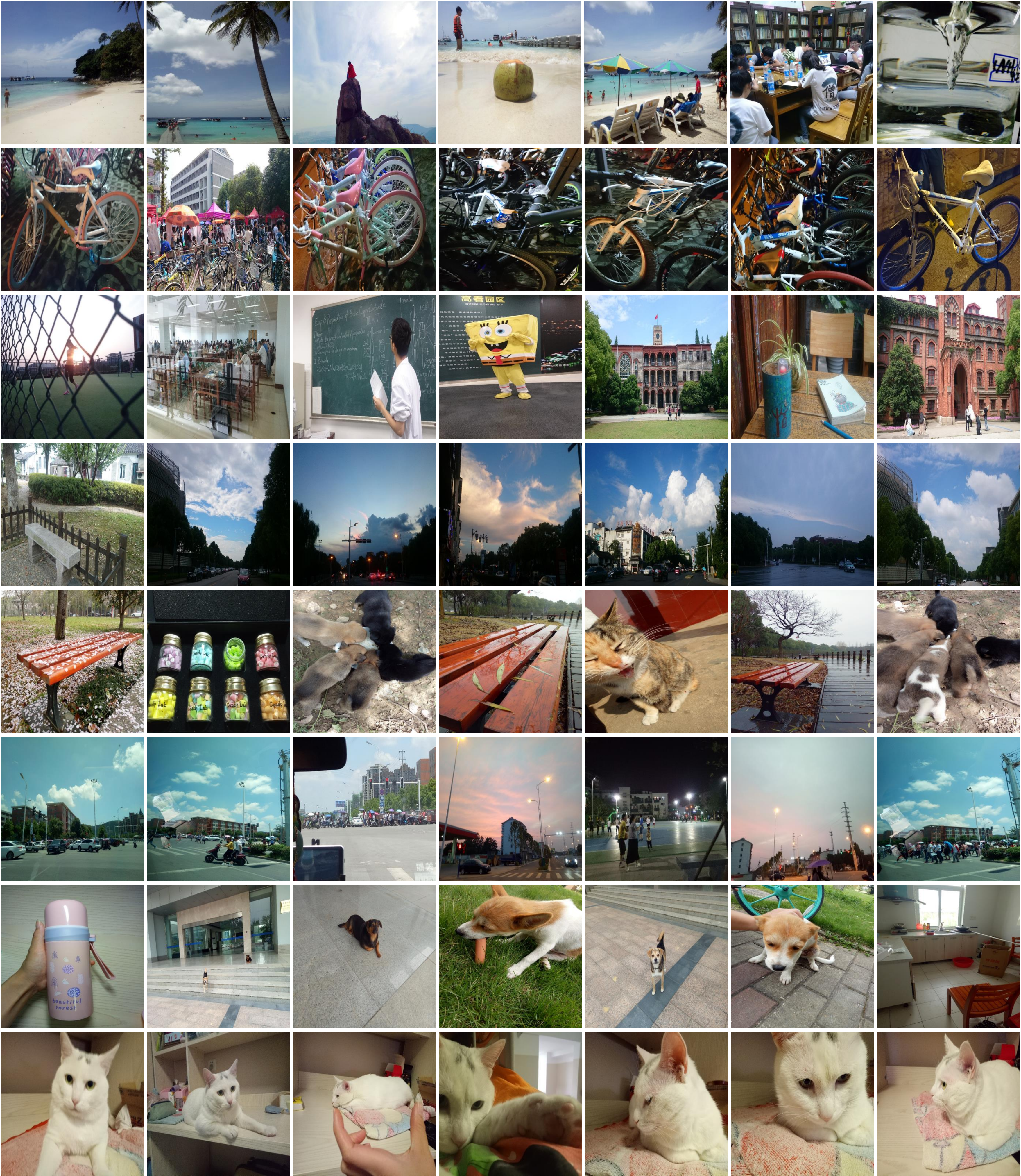} 
	\caption{Some samples from $ID1$.} \label{fig:ID1}
\end{figure*}

\begin{figure*}[t]
	\centering
	\includegraphics[width=\textwidth]{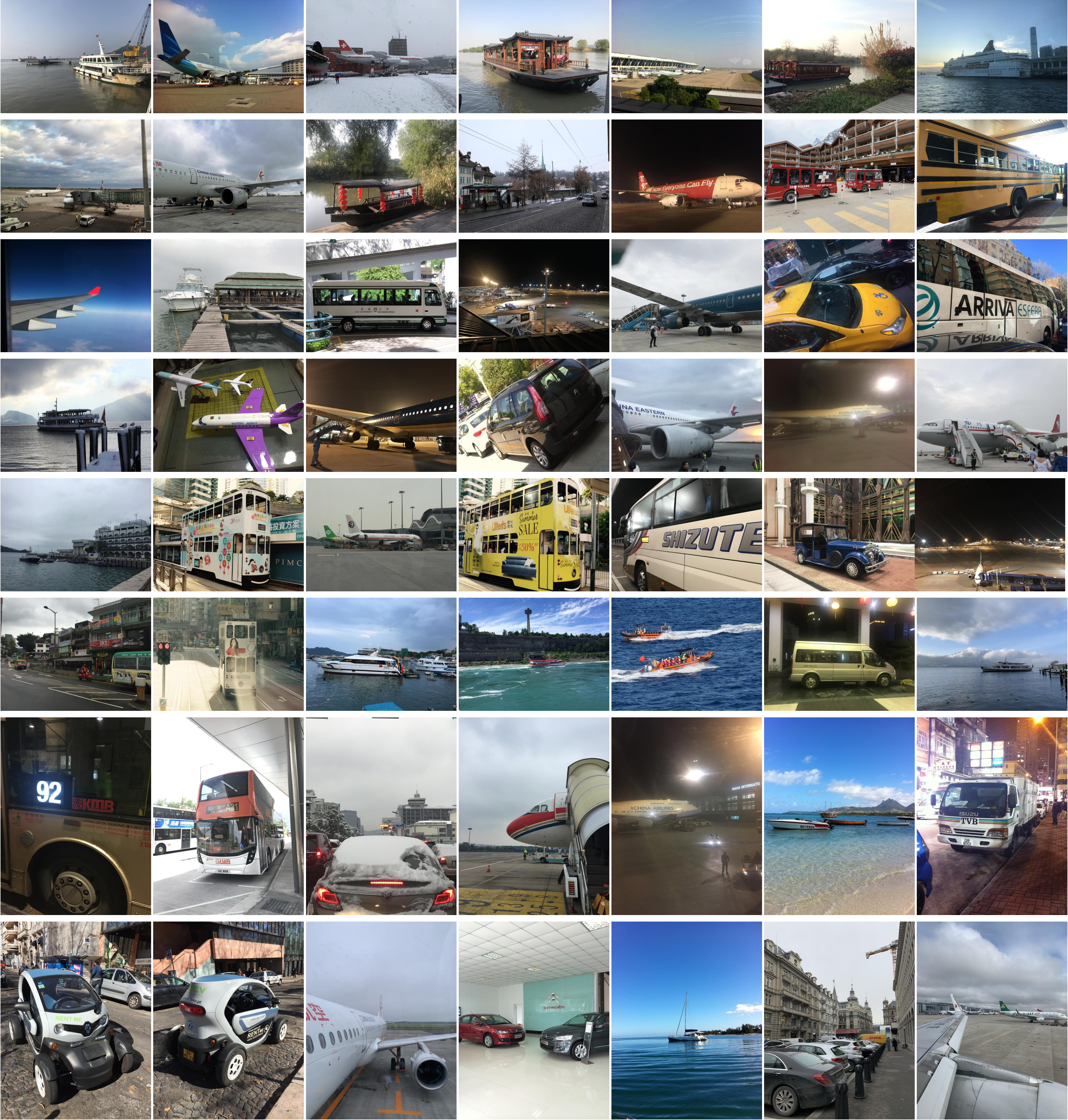} 
	\caption{Some samples from $ID15$.} \label{fig:ID15}
\end{figure*}

\textbf{High-level semantic.}
To show the high-level semantic of the presonalized images,
we illustrate examples of different users' images in 
\figref{fig:ID1}(\emph{ID1}) to \figref{fig:ID15}(\emph{ID15}),
respectively.
In Figure.2 of the main paper, we plot the proportion
of different objects different users,
which shows the inter-user gap between different user
in the perspective of object classes.
Here, we can also observe it by image examples.
By browsing the images, we can observe that a big
portion of users tend to take pictures of their daily
life.
However, since different users have different living environments, different life experiences, and different hobbies,
they tend to pay attention to different contents
in their pictures.
For example, \emph{ID7} have many images of street scenes,
\emph{ID8} have many portraits of selfie or group photos,
while \emph{ID2} and \emph{ID12} have many images of pets.
Besides that, some user also have images that seems not
very daily.
For example, \emph{ID13} took many pictures of beautiful
scenery and birds,
\emph{ID15} seems to be very interested to various means of
transportation, thus took many pictures of cars and planes.
The differences between users' images indicates the 
requirement of treating each user differently instead of
using one single model trained from public dataset.
The tendency of taking certain objects for certain user
also reflects the personalization property, \ie,
intra-user coherency of that user.

\subsection{Experiments on Urban Scene}
In this section,
we test our proposed group context module
on the adaptation of GTA5
$\xrightarrow[]{}$ Cityscapes.
GTA5 contains 24, 966 images with resolution of 1914$\times$1024.
Cityscapes contains 2, 975 training images and 500 validation images with resolution of 2048$\times$1024.
%
%
We apply mean Intersect over Union (mIoU) on the common 19
classes as the evaluation metric.
Following PLCA~\cite{PLCA} and ADVENT~\cite{vu2019advent},
we use DeepLab v2 with ResNet-101 backbone as our segmentation
network.
We set the image size the same as ADVENT\cite{vu2019advent}
in both train and test phase.
The batch size is set to 4.
We only 
conduct the first adaptation step without the second
self-training
step in this experiment,
the results are reported in ~\tabref{tab:cifar}.
Our method outperforms
the baseline approach ADVENT by 1.6\%.
Note we use a image level representation to cluster
images into groups, which might not suit
the urban images since they are all similar
urban scenes.
We believe there will be more performance
boost with more advanced cluster methods.

\subsection{Class-wise IoU}
As discussed in the main paper, our personalized image dataset
is long tail distributed.
The unbalanced class distribution may influence the MIoU
results on our dataset.
In this section, we report the class-wise IoU results
of some methods as a supplement of the MIoU results
listed in the main paper.
The results for user $ID1-ID5$ and $ID6-ID15$ 
are reported in 
\tabref{table:class_iou1} and \tabref{table:class_iou2},
respectively.
All methods use ResNet-50 as backbone.
We compare results of 4 methods in the table: 
\textit{Baseline} denotes our approach without using context;
\textit{MRNet}\cite{zheng2020unsupervised} is a recent state of
the art domain adaptation method;
\textit{OURS-S1} and \textit{OURS-S2} denote the first step
and second step of our approach, respectively.
We also report the image number of each class for all users
in the row denoted \textit{Number}.
As discussed in section 5.3 of the main paper, our proposed 
group context module can facilitate the segmentation result
for image that has other images with similar context.
On the other hand, the group context module may damage the
segmentation of rare images due to the incorporation of 
possibly irrelevant context.
Take the result of $ID6$ for example.
for
rare classes like "bicycle"(9 images), "boat"(1 image) and "bus"(2 images),
\textit{OURS-S1} obviously lag behind \textit{Baseline}.
Yet for classes like "car"(39 images), "cat"(32 images), "person"(176 images),
\textit{OURS-S1} outperforms \textit{Baseline}.
As a result, \textit{OURS-S1} only achieves 33.64 mIoU, which is lower
than 40.52 of \textit{Baseline} since the mIoU metric doesn't take the
long tailed distribution problem into consideration.
However, when we look at FIoU, \textit{OURS-S1} achieves 60.38,
which is 5.34 higher than 55.04 of \textit{Baseline}.
It is an interesting problem to investigate more sophisticated approaches
for the group context module to improve the performance on these rare
images.

\subsection{Illustration of the Clustered Groups}
\begin{figure*}[t]
	\centering
	\includegraphics[width=\textwidth]{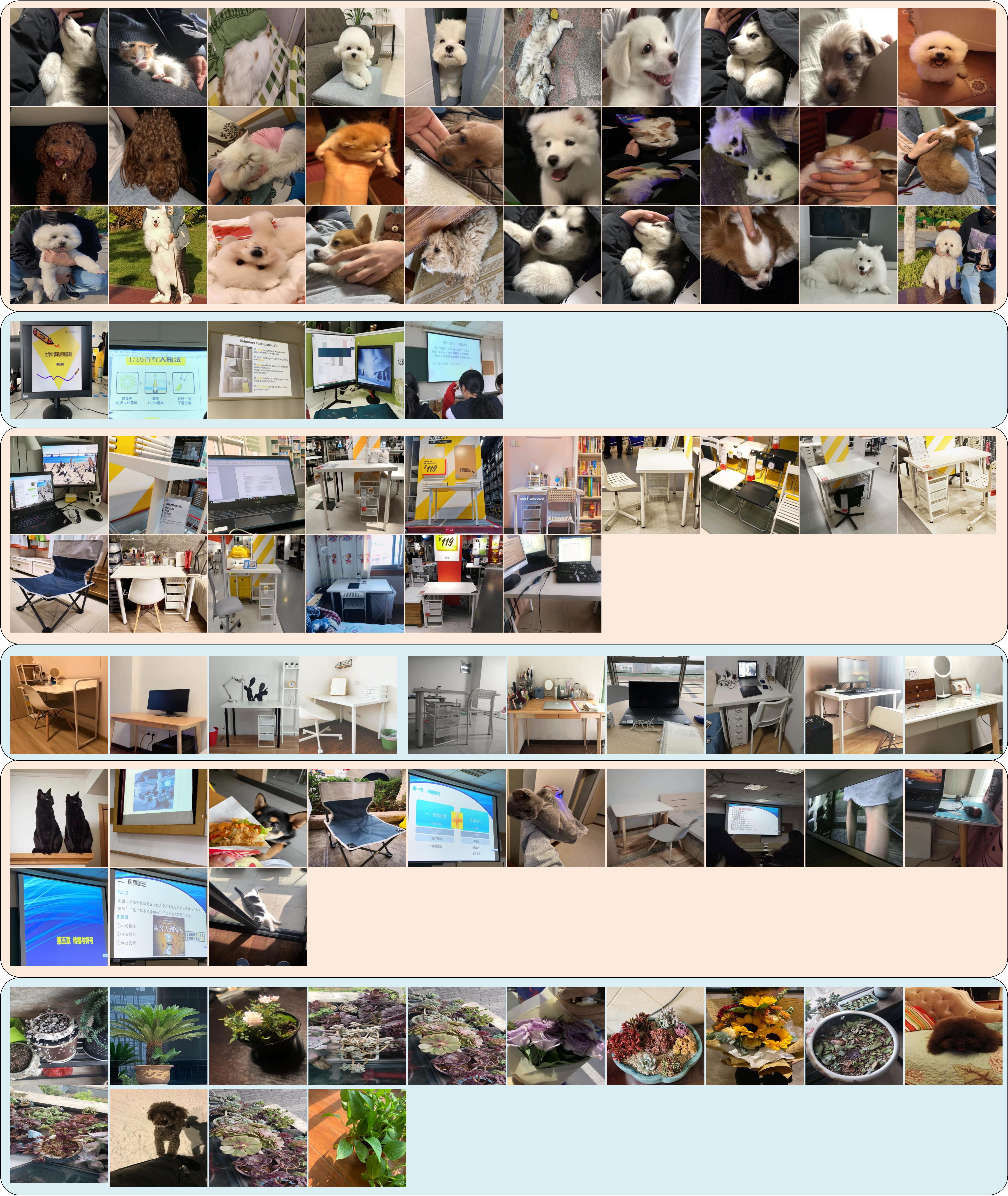} 
	\caption{Some group images from $ID12$. We cluster $ID12$ into 80 groups, and then we randomly displayed 6 of them.
	} \label{fig:group}
\end{figure*}

We display more ablation studies about clustered groups,
as shown in~\figref{fig:group}.
We use the K-means clustering algorithm to group the photos of each user.
In this section, we show some clustered groups of $ID12$ to 
give an illustrative
understanding of the group step.
As shown in \figref{fig:group}, there are many photos in the
same scene that are divided into the same group for a user.
Meanwhile, we notice there are also some undesirable cluster results.
For example, in the last group in \ref{fig:group}, there are two images of dogs while 
other images are potted plant.
The undesirable cluster results inspires us two direction to improve the system:
(1) develop more advanced cluster algorithm to get better groups, (2) improves the
group context module to utilize complementary context while discard irrelevant context.

\begin{table*}[t]
  \centering
  \renewcommand{\tabcolsep}{0.55mm} 
  
  \begin{tabular}{l|ccccccccccccccccccc|c}
    \toprule
    \multicolumn{21}{c}{GTA5 $\longrightarrow$ Cityscapes}\\
    \midrule
    Method & \rot{road} & \rot{side.} & \rot{buil.} & \rot{wall} & \rot{fence} &
    \rot{pole} & \rot{t-light} & \rot{t-sign} & \rot{vege.} & \rot{terr.} &
    \rot{sky} & \rot{pers.} & \rot{rider} & \rot{car} & \rot{truck} &
    \rot{bus} & \rot{train} & \rot{motor} & \rot{bike} & \rot{mIoU} \\
    \midrule
    Source-Only & 75.8 & 16.8 & 77.2 & 12.5 & 21.0 & 25.5 & 30.1 & 20.1 & 81.3 &
    24.6 & 70.3 & 53.8 & 26.4 & 49.9 & 17.2 & 25.9 & 6.5 & 25.3 & 36.0 & 36.6\\
    
    Fully-Supervised & -  & - & - & - & - & - & - & - & - & - & - & -  & - & - & - & - & - & - & - & 65.1 \\
    \midrule
    AdaptSeg\pub{18}~\cite{tsai2018learning} & 86.5 & 36.0 & 79.9 & 23.4 & 23.3 & 23.9 & 35.2 & 14.8 & 83.4 & 33.3 & 75.6 & 58.5 & 27.6 & 73.7 & 32.5 & 35.4 & 3.9 & 30.1 & 28.1 & 42.4 \\

    DCAN\pub{18}~\cite{dcan}         & 86.5 & 36.0 & 79.9 & 23.4 & 23.3 & 23.9 & 35.2 & 14.8 & 83.4 & 33.3 & 75.6 & 58.5 & 27.6 & 73.7 & 32.5 & 35.4 & 3.9 & 30.1 & 28.1 & 42.4 \\
    
    DISE\pub{19}~\cite{chang2019all}
                                     & 91.5 & 47.5 & 82.5 & 31.3 & 25.6 & 33.0 & 33.7 & 25.8 & 82.7 & 28.8 & 82.7 & 62.4 & 30.8 & 85.2 & 27.7 & 34.5 & 6.4 & 25.2 & 24.4 & 45.4 \\

    CLAN\pub{19}~\cite{CLAN}         & 87.0 & 27.1 & 79.6 & 27.3 & 23.3 & 28.3 & 35.5 & 24.2 & 83.6 & 27.4 & 74.2 & 58.6 & 28.0 & 76.2 & 33.1 & 36.7 & 6.7 & 31.9 & 31.4 & 43.2 \\

    ADVENT\pub{19}~\cite{vu2019advent}     & 89.9 & 36.5 & 81.6 & 29.2 & 25.2 & 28.5 & 32.3 & 22.4 & 83.9 & 34.0 & 77.1 & 57.4 & 27.9 & 83.7 & 29.4 & 39.1 & 1.5 & 28.4 & 23.3 & 43.8 \\

    SSF-DAN\pub{19}~\cite{SSFDAN}    & 90.3 & 38.9 & 81.7 & 24.8 & 22.9 & 30.5 & 37.0 & 21.2 & 84.8 & 38.8 & 76.9 & 58.8 & 30.7 & 85.7 & 30.6 & 38.1 & 5.9 & 28.3 & 36.9 & 45.4 \\

    SIBAN\pub{19}~\cite{SIBAN}       & 88.5 & 35.4 & 79.5 & 26.3 & 24.3 & 28.5 & 32.5 & 18.3 & 81.2 & 40.0 & 76.5 & 58.1 & 25.8 & 82.6 & 30.3 & 34.4 & 3.4 & 21.6 & 21.5 & 42.6 \\

    LTIR\pub{20} \textcolor{red}{w/o ST}~\cite{LTIR}         & -  & - & - & - & - & - & - & - & - & - & - & -  & - & - & - & - & - & - & - & 44.6 \\
    
    LTIR\pub{20}~\cite{LTIR}         & 92.9 & 55.0 & 85.3 & 34.2 & 31.1 & 34.9 & 40.7 & 34.0 & 85.2 & 40.1 & 87.1 & 61.0 & 31.1 & 82.5 & 32.3 & 42.9 & 0.3 & 36.4 & 46.1 & 50.2 \\

    FDA\pub{20} \textcolor{red}{w/o ST}~\cite{yang2020fda} & 90.0 & 40.5 & 79.4 & 25.3 & 26.7 & 30.6 & 31.9 & 29.3 & 79.4 & 28.8 & 76.5 & 56.4 & 27.5 & 81.7 & 27.7 & 45.1 & 17.0 & 23.8 & 29.6 & 44.6 \\

    \midrule
    Ours \textcolor{red}{w/o ST} & 89.2 & 41.5 & 83.3 & 33.3 & 15.1 & 34.6 & 42.9 & 29.0 & 85.9 &
    38.3 & 79.9 & 65.8 & 28.9 & 85.8 & 40.4 & 46.7 & 0.0 & 22.1 & 0.0 & 45.4\\
    \bottomrule
  \end{tabular}
  \caption{
    Quantitative comparison with other methods on GTA5$\xrightarrow[]{}$Cityscapes setting. \textbf{Note we don't perform the self-training step.} \textcolor{red}{ST} denotes the self-training step.
  }\label{tab:cifar}
  \vskip 0.2in
\end{table*}

\definecolor{mygray}{gray}{0.88}
\begin{table*}[htbp]
\footnotesize
	\begin{center}
		\renewcommand\tabcolsep{1.7pt}
		\begin{tabular}{l|c|ccccccccccccccccccc|l}
            \toprule[0.8pt]
            Method  & ID & \multicolumn{1}{c}{plane} & \multicolumn{1}{c}{bicycle} & \multicolumn{1}{c}{bird} & \multicolumn{1}{c}{boat} & \multicolumn{1}{c}{bottle} & \multicolumn{1}{c}{bus} & \multicolumn{1}{c}{car} & \multicolumn{1}{c}{cat} & \multicolumn{1}{c}{chair} &  \multicolumn{1}{c}{table} & \multicolumn{1}{c}{dog} & \multicolumn{1}{c}{horse} & \multicolumn{1}{c}{m.bike} & \multicolumn{1}{c}{person} & \multicolumn{1}{c}{plant} & \multicolumn{1}{c}{sheep} & \multicolumn{1}{c}{sofa} & \multicolumn{1}{c}{train} & \multicolumn{1}{c|}{tv} & \multicolumn{1}{c}{Mean} \\ \hline
            Number & \multirow{5}{*}{1} & 0 & 19 & 0 & 2 & 177 & 14 & 87 & 47 & 61 & 46 & 109 & 2 & 24 & 242 & 10 & 0 & 0 & 1 & 2 & \\ \cline{1-1}\cline{3-22}
\rowcolor{mygray} Baseline & & - & 56.06 & - & 0 & 47.16 & 46.68 & 53.94 & 66.23 & 19.9 & 31.49 & 74.5 & 29.24 & 19.9 & 52.34 & 15.91 & - & - & 27.28 & 0.01 & 36.04 \\
MRNet\cite{zheng2020unsupervised} & & - & 54.11 & - & 0.27 & 51.99 & 38.16 & 56.04 & 56.31 & 22.2 & 25.98 & 67.58 & 54.39 & 25.18 & 56.57 & 22.63 & - & - & 38.76 & 3.92 & 38.27 \\
\rowcolor{mygray} OURS-S1 & & - & 41.31 & - & 54.67 & 45.91 & 34.98 & 55.57 & 62.48 & 20.57 & 27.8 & 75.43 & 7.55 & 18.82 & 54.41 & 25.75 & - & - & 23.56 & 0.3 & 36.61 \\ 
OURS-S2 & & - & 42.86 & - & 4.68 & 42.81 & 43.51 & 56.87 & 54.29 & 17.63 & 33.6 & 69.17 & 8.58 & 8.33 & 56.48 & 19.44 & - & - & 49.5 & 0 & 33.85 \\ 
            \toprule[0.5pt]
            
            Number & \multirow{5}{*}{2} & 0 & 267 & 0 & 0 & 167 & 0 & 36 & 111 & 113 & 53 & 201 & 0 & 26 & 124 & 16 & 0 & 9 & 0 & 4 &  \\ \cline{1-1}\cline{3-22}
\rowcolor{mygray} Baseline & & - & 70.61 & - & - & 34.37 & - & 37.83 & 67.37 & 42.02 & 14.03 & 61.31 & - & 14.68 & 31.47 & 13.85 & - & 18.64 & - & 2.3 & 34.04 \\
MRNet\cite{zheng2020unsupervised} & & - & 67.14 & - & - & 50.98 & - & 38.62 & 66.95 & 26.65 & 17.9 & 64.32 & - & 14.47 & 29.75 & 24.21 & - & 14.64 & - & 4.55 & 35.02 \\
\rowcolor{mygray} OURS-S1 & & - & 66.14 & - & - & 43.46 & - & 32.19 & 67.01 & 39.9 & 15.61 & 70.58 & - & 8.85 & 35.71 & 18.33 & - & 7.95 & - & 7.47 & 34.43\\ 
OURS-S2 & & - & 72.19 & - & - & 40.84 & - & 26.68 & 69.25 & 35.15 & 14.28 & 71.97 & - & 8.24 & 37.13 & 20.3 & - & 0.77 & - & 3.74 & 33.38 \\ 
            \toprule[0.5pt]
            
            Number & \multirow{5}{*}{3} & 0 & 8 & 0 & 2 & 51 & 4 & 64 & 47 & 80 & 56 & 51 & 1 & 30 & 168 & 39 & 0 & 17 & 0 & 13 &   \\ \cline{1-1}\cline{3-22}
\rowcolor{mygray} Baseline & & - & 19.14 & - & 46.84 & 21.77 & 2.47 & 53.79 & 68.44 & 22.15 & 28.97 & 59.16 & 75.85 & 11.29 & 63.26 & 65.14 & - & 11.02 & - & 5.34 & 36.98 \\
MRNet\cite{zheng2020unsupervised} & & - & 16.13 & - & 20.14 & 27.36 & 17.71 & 60.81 & 73.04 & 25.58 & 30.75 & 59.8 & 60.22 & 24.16 & 65.34 & 60.65 & - & 9.75 & - &  3.19 & 36.98 \\
\rowcolor{mygray} OURS-S1 & & - & 4 & - & 21.16 & 11.09 & 0.88 & 68.77 & 68.18 & 28.88 & 40.03 & 57.37 & 19.43 & 17.52 & 63.19 & 56.61 & - & 10.15 & - & 10.96 & 31.88 \\ 
OURS-S2 & & - & 16.33 & - & 51.35 & 17.34 & 11.18 & 66.17 & 70.17 & 25.31 & 39.87 & 58.7 & 66.95 & 15.2 & 62.8 & 60.54 & - & 6.95 & - & 7.21 & 38.40 \\ 
            \toprule[0.5pt]
            
            Number & \multirow{5}{*}{4} & 0 & 26 & 0 & 0 & 56 & 3 & 45 & 0 & 60 & 48 & 0 & 0 & 23 & 129 & 41 & 0 & 10 & 0 & 9 &   \\ \cline{1-1}\cline{3-22}
\rowcolor{mygray} Baseline & & - & 68.98 & - & - & 53.17 & 31.65 & 61.62 & - & 30.23 & 24.49 & - & - & 11.78 & 70.51 & 52.07 & - & 35.33 & - & 0 & 39.98 \\
MRNet\cite{zheng2020unsupervised} & & - & 60.01 & - & - & 60.14 & 28.21 & 69.56 & - & 31.04 & 9.93 & - & - & 10.67 & 74.17 & 51.86 & - & 6.33 & - & 0 & 36.54 \\
\rowcolor{mygray} OURS-S1 & & - & 66.28 & - & - & 56.95 & 24.89 & 73.41 & - & 33.62 & 23.62 & - & - & 6.13 & 75.21 & 53.54 & - & 30.36 & - & 0 & 40.36 \\ 
OURS-S2 & & - & 55.5 & - & - & 53.53 & 23.07 & 79.65 & - & 31.29 & 33.58 & - & - & 5.92 & 72.71 & 62.36 & - & 37.22 & - & 0.11 & 41.36 \\ 
            \toprule[0.5pt]
            
            Number & \multirow{5}{*}{5} & 0 & 36 & 30 & 1 & 26 & 5 & 54 & 76 & 22 & 25 & 31 & 0 & 17 & 74 & 4 & 0 & 3 & 28 & 23 &    \\ \cline{1-1}\cline{3-22}
\rowcolor{mygray} Baseline & & - & 69 & 72.52 & 0 & 52.03 & 75.36 & 81.65 & 78.82 & 2.92 & 21.65 & 59.49 & - & 15.38 & 36.36 & 4.24 & - & 0 & 75.65 & 55.1 & 43.76 \\
MRNet\cite{zheng2020unsupervised} & & - & 62.29 & 70.63 & 0 & 47.89 & 73.75 & 80.82 & 80.36 & 0.78 & 7.85 & 65.37 & - & 27.9 & 48.78 & 0.86 & - & 20.71 & 65.48 & 50.41 & 43.99 \\
\rowcolor{mygray} OURS-S1 & & - & 66.77 & 59.21 & 0 & 37.47 & 78.14 & 84.88 & 77.17 & 2.96 & 19.7 & 64.23 & - & 21.17 & 44.54 & 12.77 & - & 0 & 73.67 & 65.26 & 44.25 \\ 
OURS-S2 & & - & 71.56 & 69.59 & 0 & 41.27 & 82.77 & 80.12 & 79.03 & 1.28 & 35.9 & 68.19 & - & 22.24 & 54.55 & 6.34 & - & 0 & 75.91 & 58.86 & 46.72 \\

            \toprule[0.8pt]
        \end{tabular}
	\end{center}
	\vspace{-5pt}
	\caption{Class-wise IoU results of different
	methods on user $ID1-5$.
	The row "number" indicates how many number of images contains certain class of object
	in the test set.
	}
	\label{table:class_iou1}
\end{table*}

\begin{table*}[htbp]
\footnotesize
	\begin{center}
		\renewcommand\tabcolsep{1.7pt}
		\begin{tabular}{l|c|ccccccccccccccccccc|l}
            \toprule[0.8pt]
            Method  & ID & \multicolumn{1}{c}{plane} & \multicolumn{1}{c}{bicycle} & \multicolumn{1}{c}{bird} & \multicolumn{1}{c}{boat} & \multicolumn{1}{c}{bottle} & \multicolumn{1}{c}{bus} & \multicolumn{1}{c}{car} & \multicolumn{1}{c}{cat} & \multicolumn{1}{c}{chair} &  \multicolumn{1}{c}{table} & \multicolumn{1}{c}{dog} & \multicolumn{1}{c}{horse} & \multicolumn{1}{c}{m.bike} & \multicolumn{1}{c}{person} & \multicolumn{1}{c}{plant} & \multicolumn{1}{c}{sheep} & \multicolumn{1}{c}{sofa} & \multicolumn{1}{c}{train} & \multicolumn{1}{c|}{tv} & \multicolumn{1}{c}{Mean} \\ \hline
  
            Number & \multirow{5}{*}{6} & 0 & 9 & 0 & 1 & 66 & 2 & 39 & 32 & 71 & 36 & 60 & 0 & 9 & 176 & 34 & 0 & 13 & 0 & 4 &     \\ \cline{1-1}\cline{3-22}
\rowcolor{mygray} Baseline & & - & 56.93 & - & 22.87 & 47.74 & 45.37 & 31.27 & 77.46 & 14.77 & 32.74 & 69.24 & - & 25.24 & 60.78 & 66.24 & - & 12.58 & - & 4.09 & 40.52 \\
MRNet\cite{zheng2020unsupervised} & & - & 61.65 & - & 0 & 51.78 & 61.25 & 32.99 & 73.44 & 24.24 & 32.75 & 71.01 & - & 17.58 & 66.11 & 62.36 & - & 16.77 & - & 0.71 & 40.90 \\
\rowcolor{mygray} OURS-S1 & & - & 6.77 & - & 19.94 & 46.33 & 7.11 & 36.45 & 78.14 & 21.04 & 30.67 & 68.52 & - & 9.09 & 65 & 65.15 & - & 7.64 & - & 9.09 & 33.64 \\ 
OURS-S2 & & - & 36.89 & - & 21.09 & 31.26 & 51.77 & 40.24 & 68.66 & 24.38 & 37.34 & 69.9 & - & 7.86 & 65.65 & 58.63 & - & 10.43 & - & 2.04 & 37.58 \\ 
            \toprule[0.5pt]
            
            Number & \multirow{5}{*}{7} & 0 & 45 & 26 & 0 & 28 & 16 & 80 & 28 & 27 & 13 & 6 & 0 & 61 & 166 & 18 & 0 & 0 & 0 & 4 &      \\ \cline{1-1}\cline{3-22}
\rowcolor{mygray} Baseline & & - & 40.05 & 64.25 & - & 0.61 & 26.8 & 77.09 & 81.7 & 33.16 & 19.19 & 22.13 & - & 58.27 & 77.13 & 40.35 & - & - & - & 0 & 41.59 \\
MRNet\cite{zheng2020unsupervised} & & - & 49.83 & 54.09 & - & 1.75 & 26.31 & 76.96 & 88.05 & 20.58 & 16.42 & 8.78 & - & 58 & 80.32 & 41.74 & - & - & - & 0 & 40.22 \\
\rowcolor{mygray} OURS-S1 & & - & 49.05 & 44.63 & - & 2.64 & 38.72 & 74.38 & 81.54 & 23.83 & 3.68 & 8.44 & - & 57.46 & 77.7 & 33.81 & - & - & - & 0 & 38.14 \\ 
OURS-S2 & & - & 54.21 & 62.57 & - & 4.7 & 50.44 & 79.03 & 87.74 & 39.68 & 10.2 & 6.23 & - & 59.55 & 78.59 & 41.49 & - & - & - & 0 & 44.19 \\ 
            \toprule[0.5pt]
            
            Number & \multirow{5}{*}{8} & 0 & 6 & 0 & 13 & 34 & 3 & 39 & 76 & 60 & 59 & 41 & 0 & 10 & 201 & 18 & 0 & 19 & 0 & 9 &      \\ \cline{1-1}\cline{3-22}
\rowcolor{mygray} Baseline & & - & 15.93 & - & 18.89 & 31.23 & 4.17 & 54.33 & 74.94 & 6.61 & 19.04 & 56.13 & - & 8.73 & 73.59 & 26.33 & - & 17.84 & - & 7.93 & 29.69 \\
MRNet\cite{zheng2020unsupervised} & & - & 19.44 & - & 16.21 & 35.69 & 63.37 & 56.97 & 75.43 & 11.13 & 26.01 & 63.51 & - & 14.2 & 76.62 & 24.81 & - & 6.97 & - & 17.28 & 36.26 \\
\rowcolor{mygray} OURS-S1 & & - & 9.25 & - & 15.21 & 44.85 & 17.84 & 58.45 & 77.29 & 9.95 & 26.09 & 59.67 & - & 3.47 & 80.78 & 21.29 & - & 17.79 & - & 9.63 & 32.25 \\ 
OURS-S2 & & - & 9.39 & - & 32.35 & 47.81 & 24.01 & 58.36 & 79.15 & 6.96 & 8.3 & 65.49 & - & 25.14 & 80.7 & 22.98 & - & 32.52 & - & 23.06 & 36.87 \\ 
            \toprule[0.5pt]
            
            Number & \multirow{5}{*}{9} & 0 & 69 & 1 & 0 & 42 & 5 & 85 & 2 & 88 & 39 & 24 & 0 & 45 & 85 & 31 & 0 & 5 & 0 & 0 &      \\ \cline{1-1}\cline{3-22}
\rowcolor{mygray} Baseline & & - & 62.81 & 22.83 & - & 42.13 & 0 & 87.76 & 5.32 & 24.81 & 8.28 & 78.95 & - & 9.23 & 65.4 & 10.22 & - & 53.65 & - & - & 36.26 \\
MRNet\cite{zheng2020unsupervised} & & - & 64.65 & 0.7 & - & 53.46 & 4.05 & 85.41 & 1.98 & 37.43 & 3.79 & 67.21 & - & 12.15 & 58.87 & 19.68 & - & 11.2 & - & - & 32.35 \\
\rowcolor{mygray} OURS-S1 & & - & 58.23 & 7.02 & - & 53.64 & 26.56 & 90.85 & 2.99 & 31.31 & 14.16 & 72.39 & - & 33.5 & 65.58 & 16.25 & - & 45.78 & - & - & 39.87 \\ 
OURS-S2 & & - & 64.74 & 28.45 & - & 55.6 & 64.46 & 92.29 & 2.29 & 27.02 & 12.86 & 68.07 & - & 43.29 & 63.89 & 12.35 & - & 45.21 & - & - & 44.66 \\ 
            \toprule[0.5pt]
            
            Number & \multirow{5}{*}{10} & 0 & 19 & 0 & 0 & 25 & 1 & 20 & 0 & 40 & 35 & 1 & 0 & 5 & 59 & 36 & 0 & 4 & 0 & 22 &       \\ \cline{1-1}\cline{3-22}
\rowcolor{mygray} Baseline & & - & 71.64 & - & - & 65.33 & 0 & 80.5 & - & 24.21 & 20.07 & 0 & - & 17.59 & 62.95 & 52.09 & - & 13 & - & 62.95 & 39.19 \\
MRNet\cite{zheng2020unsupervised} & & - & 65.25 & - & - & 47.52 & 0 & 71.5 & - & 26.94 & 13.51 & 0 & - & 8.94 & 60.74 & 62.26 & - & 6.88 & - & 33.71 & 33.10 \\
\rowcolor{mygray} OURS-S1 & & - & 68.72 & - & - & 66.3 & 0 & 72.79 & - & 32.85 & 23.02 & 0 & - & 15.7 & 66.74 & 62.45 & - & 0 & - & 55.66 & 38.69 \\ 
OURS-S2 & & - & 71.69 & - & - & 60.39 & 0 & 73.14 & - & 32.41 & 23.42 & 0 & - & 13.16 & 64.39 & 64.53 & - & 43.94 & - & 57.29 & 42.03 \\ 
            \toprule[0.5pt]

            Number & \multirow{5}{*}{11} & 0 & 4 & 0 & 0 & 18 & 2 & 28 & 52 & 3 & 1 & 16 & 0 & 9 & 69 & 6 & 1 & 3 & 0 & 7 &        \\ \cline{1-1}\cline{3-22}
\rowcolor{mygray} Baseline & & - & 11.23 & - & - & 47.83 & 21.54 & 62.95 & 71.66 & 13.5 & 10.6 & 17.56 & - & 49.02 & 71.81 & 9.81 & 43.8 & 0.51 & - & 36.68 & 33.46 \\
MRNet\cite{zheng2020unsupervised} & & - & 16.46 & - & - & 47.56 & 84.47 & 59.19 & 76.77 & 15.65 & 0.4 & 32.23 & - & 58.21 & 67.63 & 7.2 & 0 & 0.39 & - & 41.53 & 36.26 \\
\rowcolor{mygray} OURS-S1 & & - & 28.98 & - & - & 43.9 & 59.65 & 69.36 & 79.54 & 10.05 & 0 & 33.22 & - & 42.17 & 75.35 & 16.39 & 52.85 & 0 & - & 9.35 & 37.20 \\ 
OURS-S2 & & - & 24.91 & - & - & 49.03 & 68.91 & 72.16 & 84.57 & 22.72 & 0 & 59.94 & - & 46.26 & 74.86 & 13.59 & 0 & 0 & - & 6.88 & 37.42 \\ 
            \toprule[0.5pt]
            
            Number & \multirow{5}{*}{12} & 0 & 0 & 0 & 0 & 9 & 0 & 1 & 39 & 31 & 38 & 37 & 0 & 1 & 30 & 23 & 0 & 1 & 0 & 9 &         \\ \cline{1-1}\cline{3-22}
\rowcolor{mygray} Baseline & & - & - & - & - & 6.01 & - & 6.39 & 74.42 & 40.52 & 17.78 & 74.03 & - & 65.16 & 25.14 & 78.13 & - & 0 & - & 42.02 & 39.05 \\
MRNet\cite{zheng2020unsupervised} & & - & - & - & - & 6.38 & - & 1.14 & 64.38 & 37.19 & 2.71 & 59.5 & - & 16.17 & 36.53 & 75.41 & - & 0 & - & 50.17 & 31.78 \\
\rowcolor{mygray} OURS-S1 & & - & - & - & - & 6.63 & - & 50.24 & 79.43 & 44.4 & 21.21 & 79.72 & - & 68.67 & 22.54 & 73.7 & - & 0 & - & 20.32 & 42.44 \\ 
OURS-S2 & & - & - & - & - & 4.9 & - & 47.66 & 66.57 & 49.15 & 24.95 & 60.64 & - & 50.92 & 39.78 & 78.04 & - & 0.41 & - & 57.82 & 43.71 \\ 
            \toprule[0.5pt]
            
            Number & \multirow{5}{*}{13} & 2 & 4 & 56 & 22 & 3 & 0 & 9 & 0 & 9 & 6 & 0 & 0 & 0 & 65 & 2 & 0 & 0 & 0 & 0 &          \\ \cline{1-1}\cline{3-22}
\rowcolor{mygray} Baseline & & 52.85 & 60.03 & 67.04 & 48.67 & 0 & - & 28.98 & - & 17.88 & 23.66 & - & - & - & 77.91 & 4.7 & - & - & - & - & 38.17 \\
MRNet\cite{zheng2020unsupervised} & & 71.6 & 59.41 & 64.05 & 41.13 & 0 & - & 22.49 & - & 19.71 & 20.73 & - & - & - & 74.22 & 4.39 & - & - & - & - & 37.77 \\
\rowcolor{mygray} OURS-S1 & & 62.55 & 52.58 & 70.59 & 47.18 & 1.64 & - & 19.61 & - & 17.79 & 36.6 & - & - & - & 75.19 & 12.27 & - & - & - & - & 39.6 \\ 
OURS-S2 & & 46.76 & 55.86 & 71.14 & 39.02 & 0 & - & 20.91 & - & 11.13 & 8.77 & - & - & - & 79.77 & 17.8 & - & - & - & - & 35.12 \\ 
            \toprule[0.5pt]
            
            Number & \multirow{5}{*}{14} & 1 & 37 & 0 & 1 & 21 & 2 & 61 & 22 & 36 & 26 & 26 & 0 & 21 & 70 & 18 & 0 & 3 & 0 & 13 &           \\ \cline{1-1}\cline{3-22}
\rowcolor{mygray} Baseline & & 42.72 & 59.65 & - & 0 & 55.48 & 2.74 & 74.65 & 61.56 & 13.53 & 11.76 & 47.49 & - & 13.35 & 54.66 & 47.02 & - & 0.75 & - & 16.11 & 33.43 \\
MRNet\cite{zheng2020unsupervised} & & 49.98 & 58.48 & - & 0 & 49.67 & 0 & 68.03 & 80.52 & 17.29 & 18.81 & 50.64 & - & 13.76 & 51.1 & 57.08 & - & 5.29 & - & 17.67 & 35.89 \\
\rowcolor{mygray} OURS-S1 & & 11.46 & 52.89 & - & 0 & 19.73 & 0 & 74.82 & 66.82 & 19.46 & 8.41 & 57.03 & - & 10.26 & 58.39 & 60.8 & - & 0 & - & 15.46 & 30.37 \\ 
OURS-S2 & & 37.14 & 60.44 & - & 0 & 13.61 & 0.33 & 75.78 & 74.38 & 23.9 & 20.11 & 60.77 & - & 23.29 & 51.46 & 59.52 & - & 0 & - & 11.99 & 34.18 \\ 
            \toprule[0.5pt]
            
            Number & \multirow{5}{*}{15} & 26 & 1 & 0 & 22 & 30 & 8 & 44 & 0 & 26 & 35 & 0 & 0 & 3 & 48 & 5 & 0 & 12 & 2 & 6 &           \\ \cline{1-1}\cline{3-22}
\rowcolor{mygray} Baseline & & 66.49 & 5.35 & - & 58.21 & 63.3 & 55.34 & 69.83 & - & 37.96 & 14.34 & - & - & 2.52 & 44.7 & 25.34 & - & 46.76 & 4.83 & 29.2 & 37.44 \\
MRNet\cite{zheng2020unsupervised} & & 62.6 & 6.72 & - & 58.17 & 68.13 & 54.23 & 75.39 & - & 29.39 & 21.46 & - & - & 0 & 13.51 & 27.88 & - & 25.78 & 5.82 & 2.34 & 32.24 \\
\rowcolor{mygray} OURS-S1 & & 69.83 & 23.87 & - & 59.73 & 65.62 & 67.73 & 77.27 & - & 33.13 & 23.44 & - & - & 0.57 & 29.28 & 39.47 & - & 36.72 & 47.34 & 16.55 & 42.18 \\ 
OURS-S2 & & 74.6 & 0.00 & - & 66.01 & 60.3 & 63.89 & 71.69 & - & 33.21 & 23.7 & - & - & 0 & 23.25 & 39.98 & - & 38.24 & 3 & 32.55 & 37.89 \\ 

            \toprule[0.8pt]
        \end{tabular}
	\end{center}
	\vspace{-10pt}
	\caption{Class-wise IoU results of different
	methods on user $ID6-15$.
	The row "number" indicates how many number of images contains certain class of object
	in the test set.
	}
	\label{table:class_iou2}
\end{table*}

\begin{figure*}[t]
	\centering
	\includegraphics[width=\textwidth]{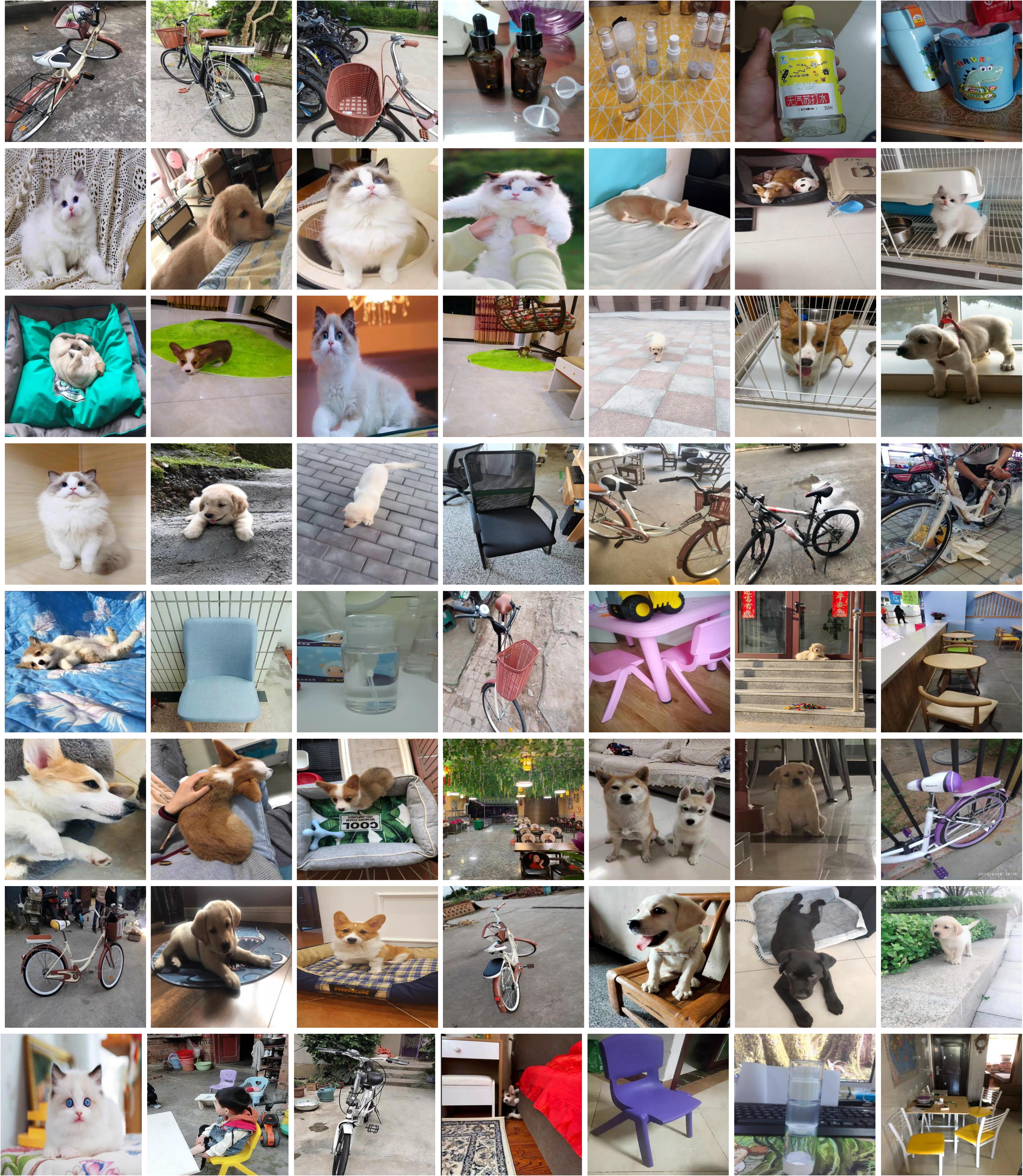} 
	\caption{Some samples from $ID2$.} \label{fig:ID2}
\end{figure*}

\begin{figure*}[t]
	\centering
	\includegraphics[width=\textwidth]{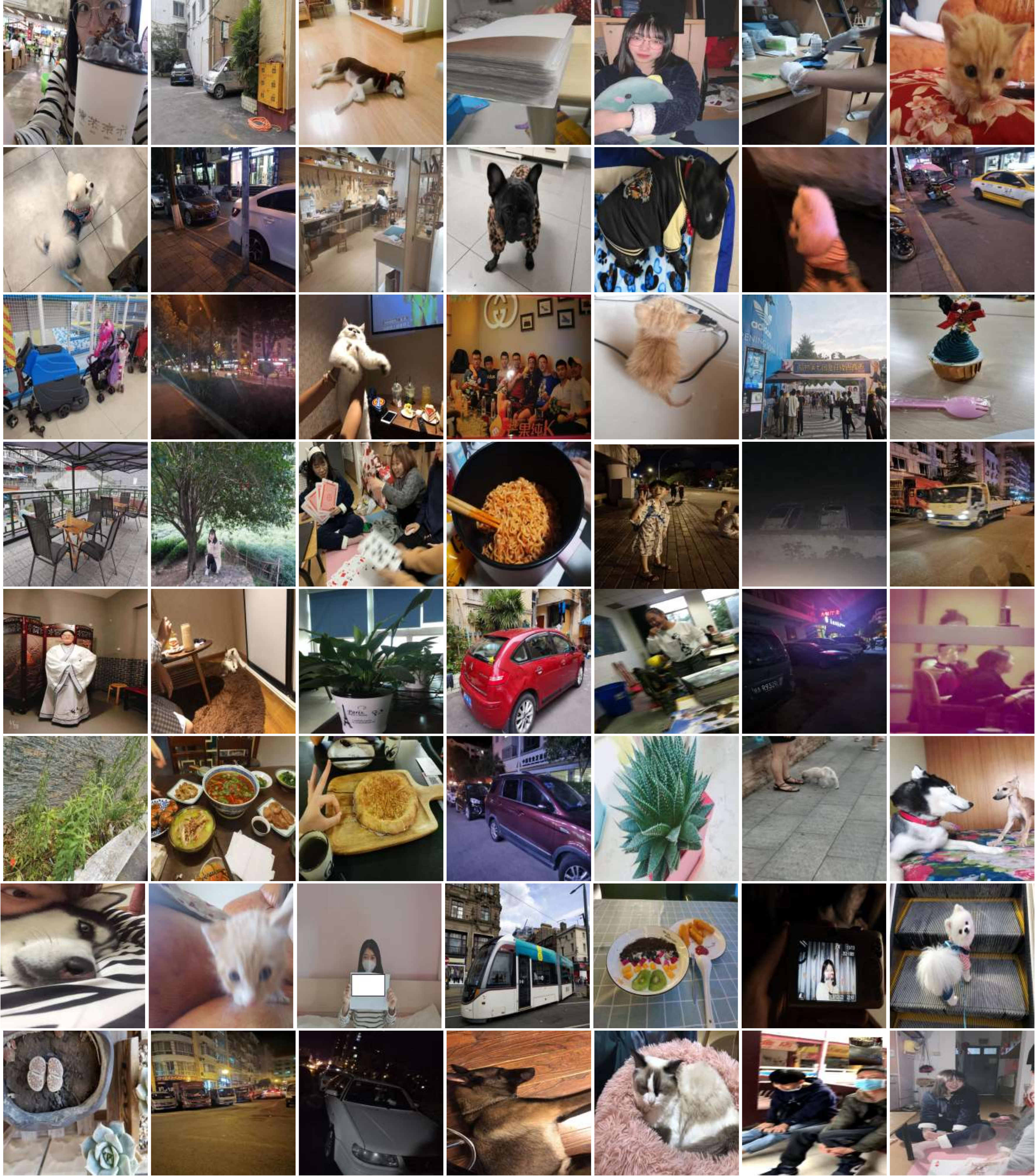} 
	\caption{Some samples from $ID3$.} \label{fig:ID3}
\end{figure*}

\begin{figure*}[t]
	\centering
	\includegraphics[width=\textwidth]{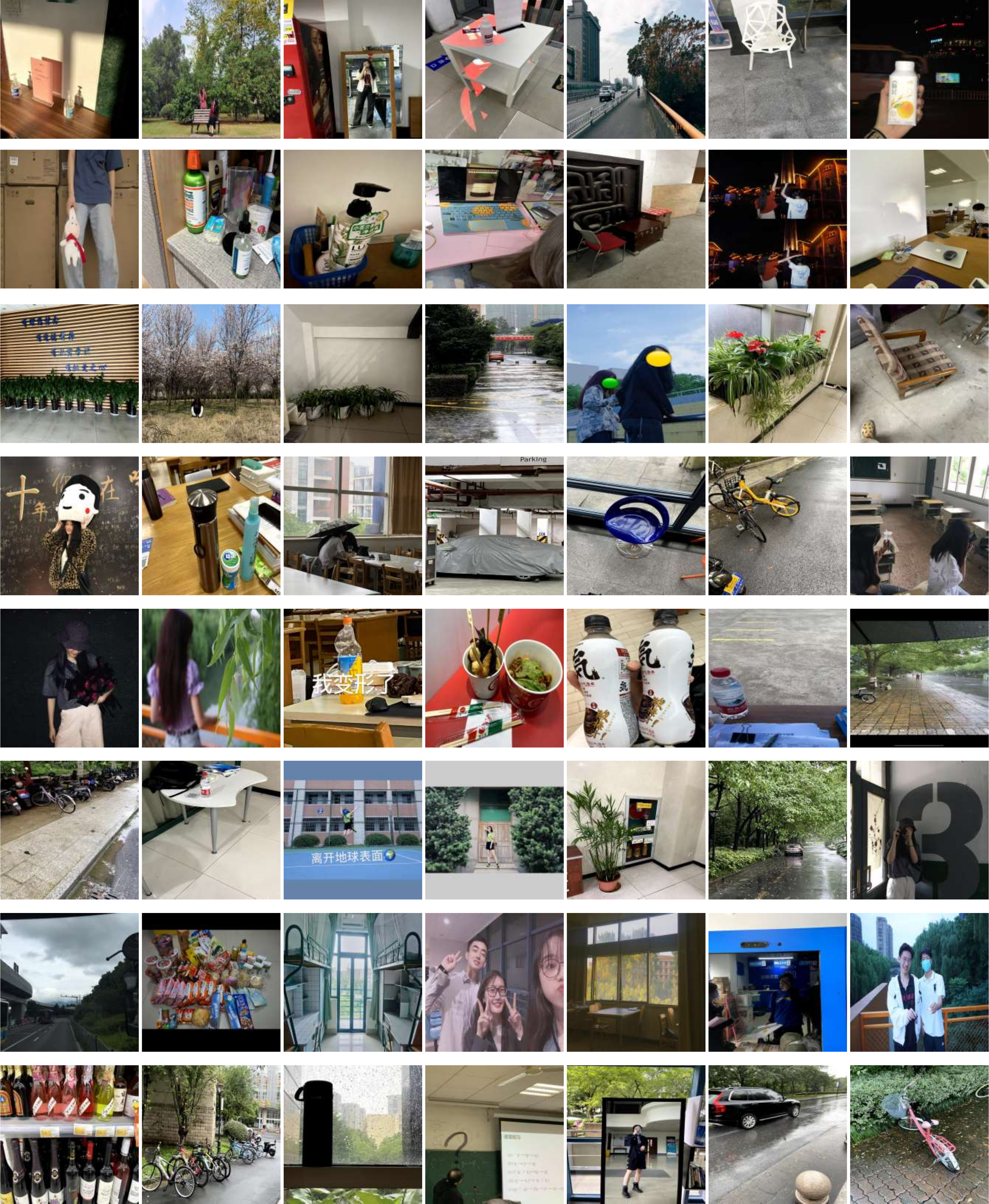} 
	\caption{Some samples from $ID4$.} \label{fig:ID4}
\end{figure*}

\begin{figure*}[t]
	\centering
	\includegraphics[width=\textwidth]{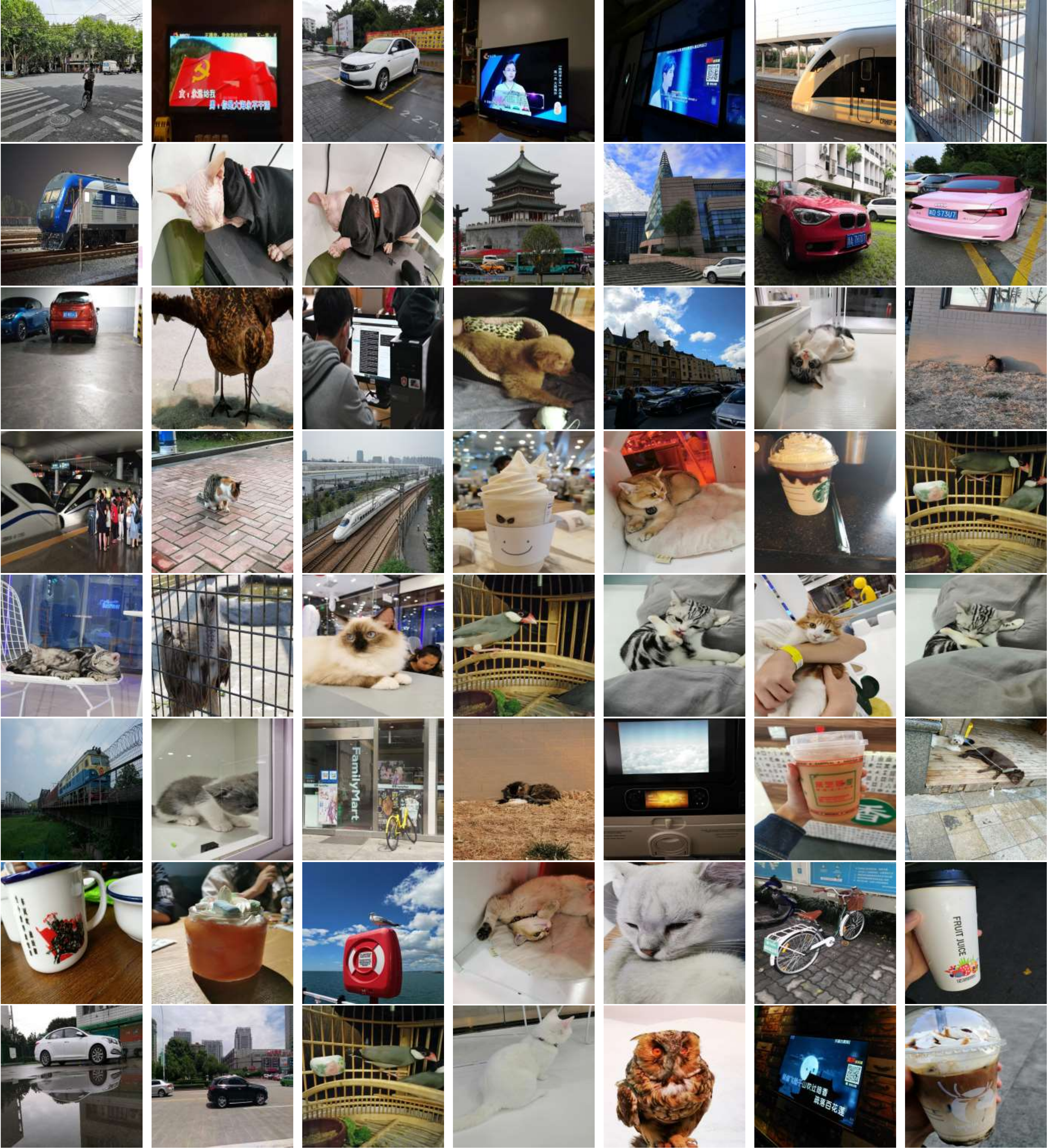} 
	\caption{Some samples from $ID5$.} \label{fig:ID5}
\end{figure*}

\begin{figure*}[t]
	\centering
	\includegraphics[width=\textwidth]{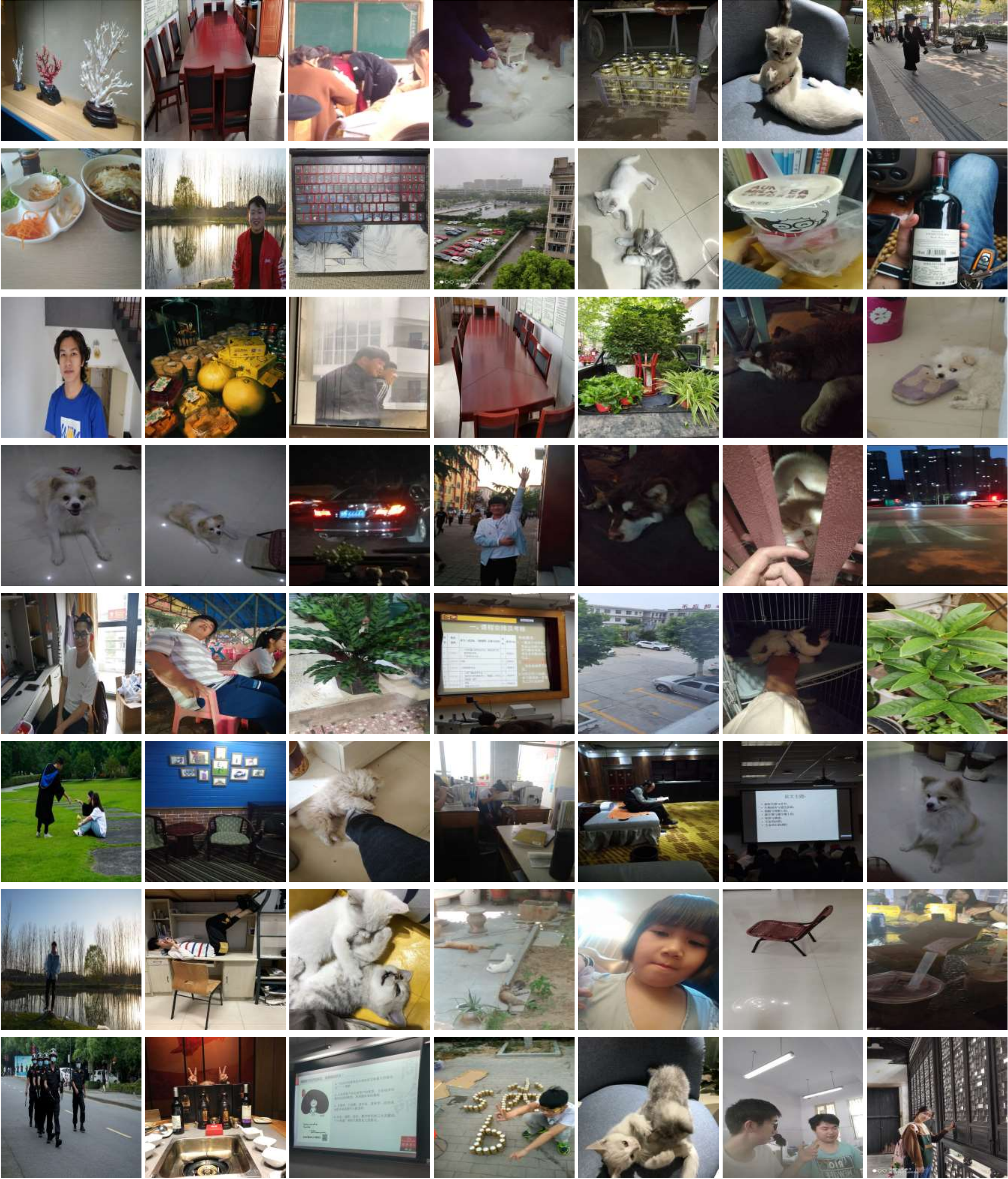} 
	\caption{Some samples from $ID6$.} \label{fig:ID6}
\end{figure*}

\begin{figure*}[t]
	\centering
	\includegraphics[width=\textwidth]{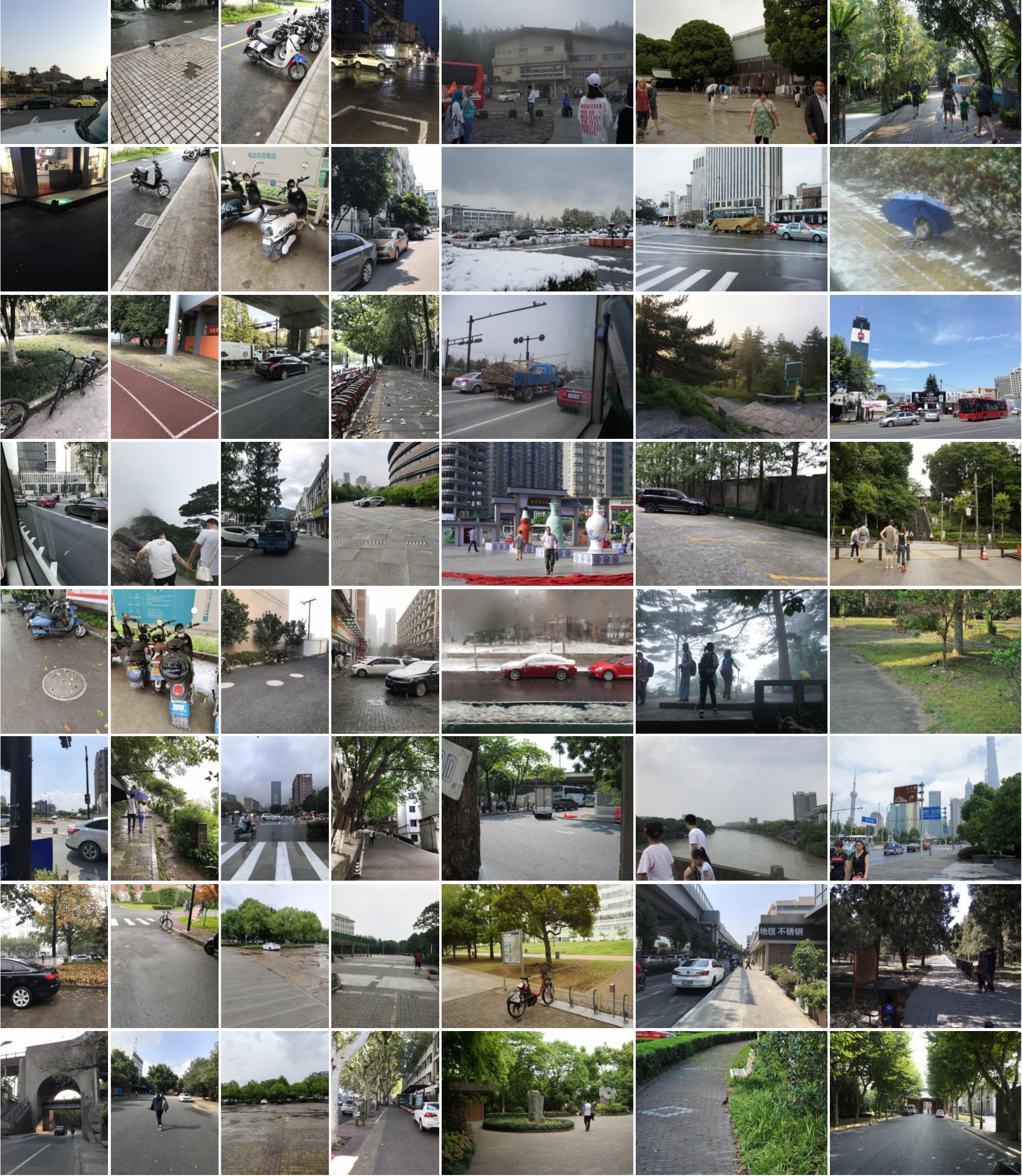} 
	\caption{Some samples from $ID7$.} \label{fig:ID7}
\end{figure*}

\begin{figure*}[t]
	\centering
	\includegraphics[width=\textwidth]{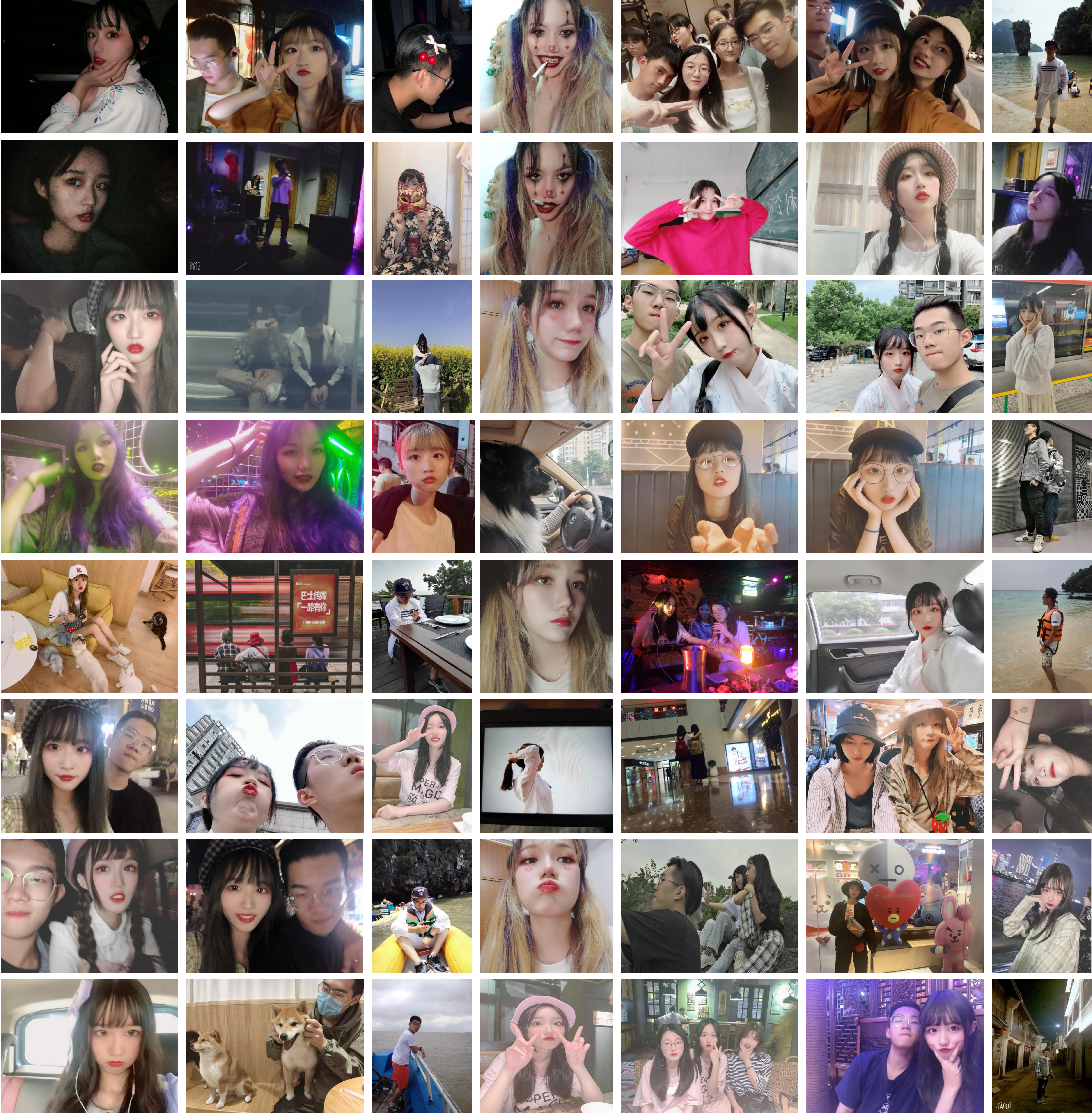} 
	\caption{Some samples from $ID8$.} \label{fig:ID8}
\end{figure*}

\begin{figure*}[t]
	\centering
	\includegraphics[width=\textwidth]{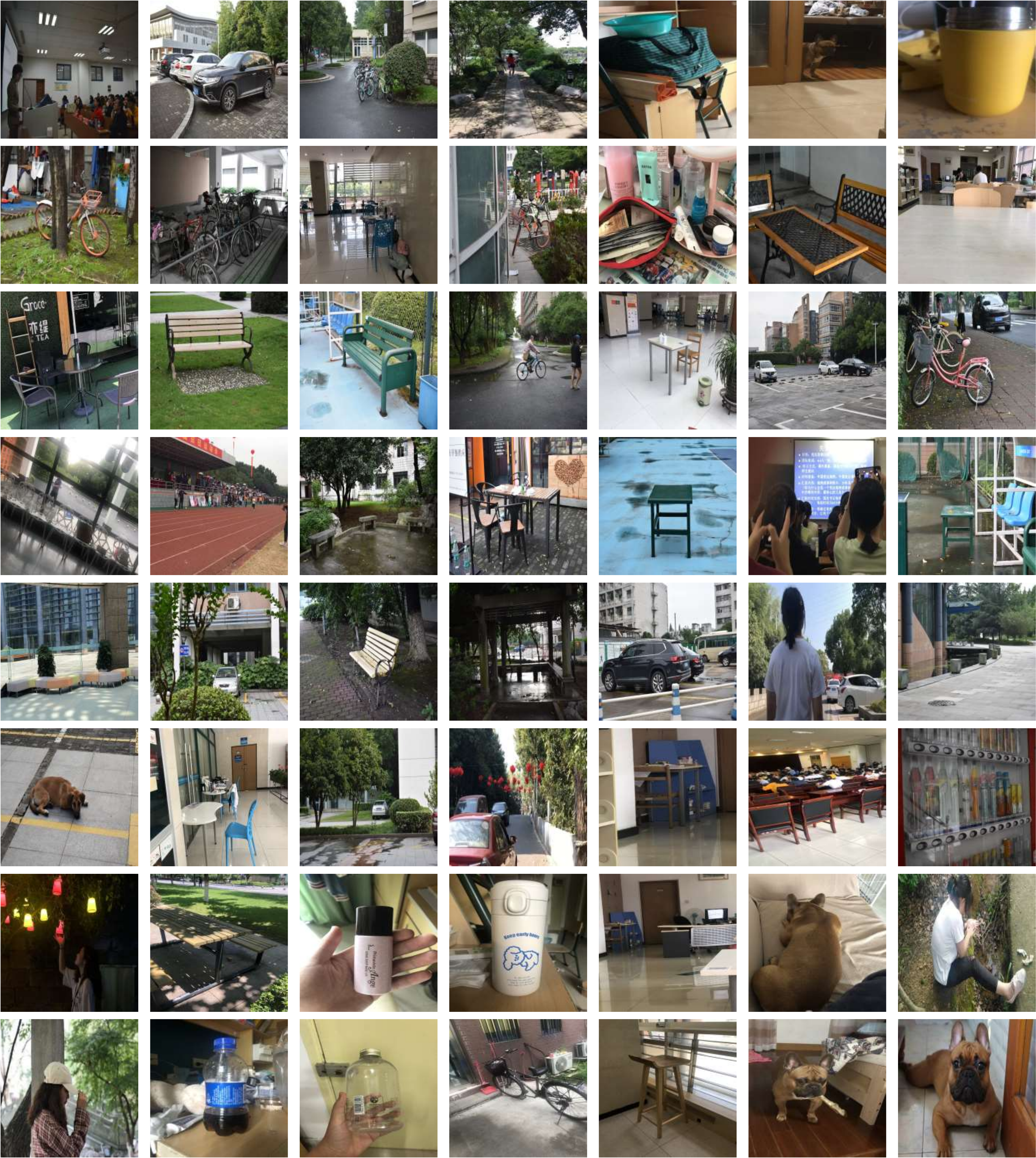} 
	\caption{Some samples from $ID9$.} \label{fig:ID9}
\end{figure*}

\begin{figure*}[t]
	\centering
	\includegraphics[width=\textwidth]{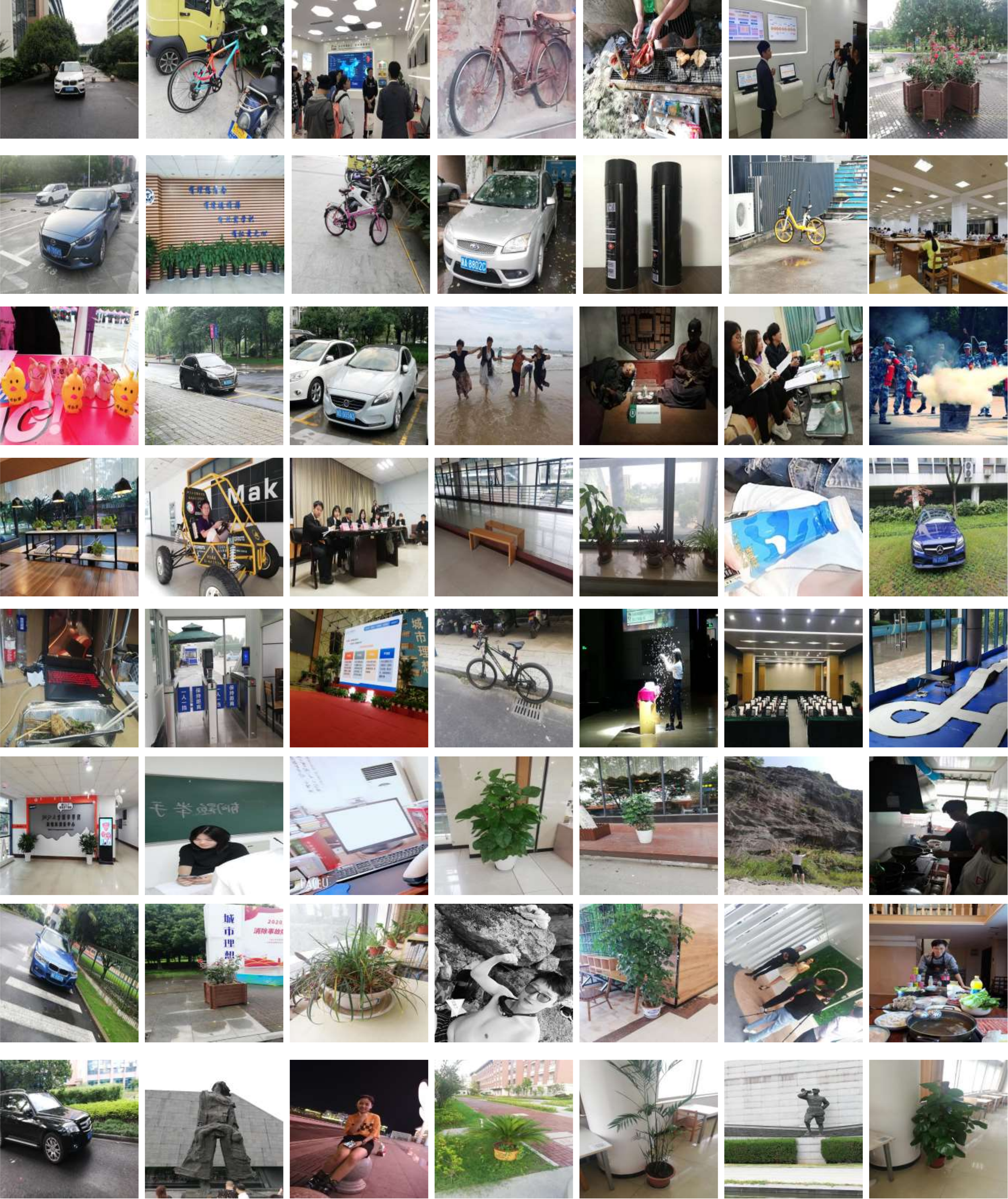} 
	\caption{Some samples from $ID10$.} \label{fig:ID10}
\end{figure*}

\begin{figure*}[t]
	\centering
	\includegraphics[width=\textwidth]{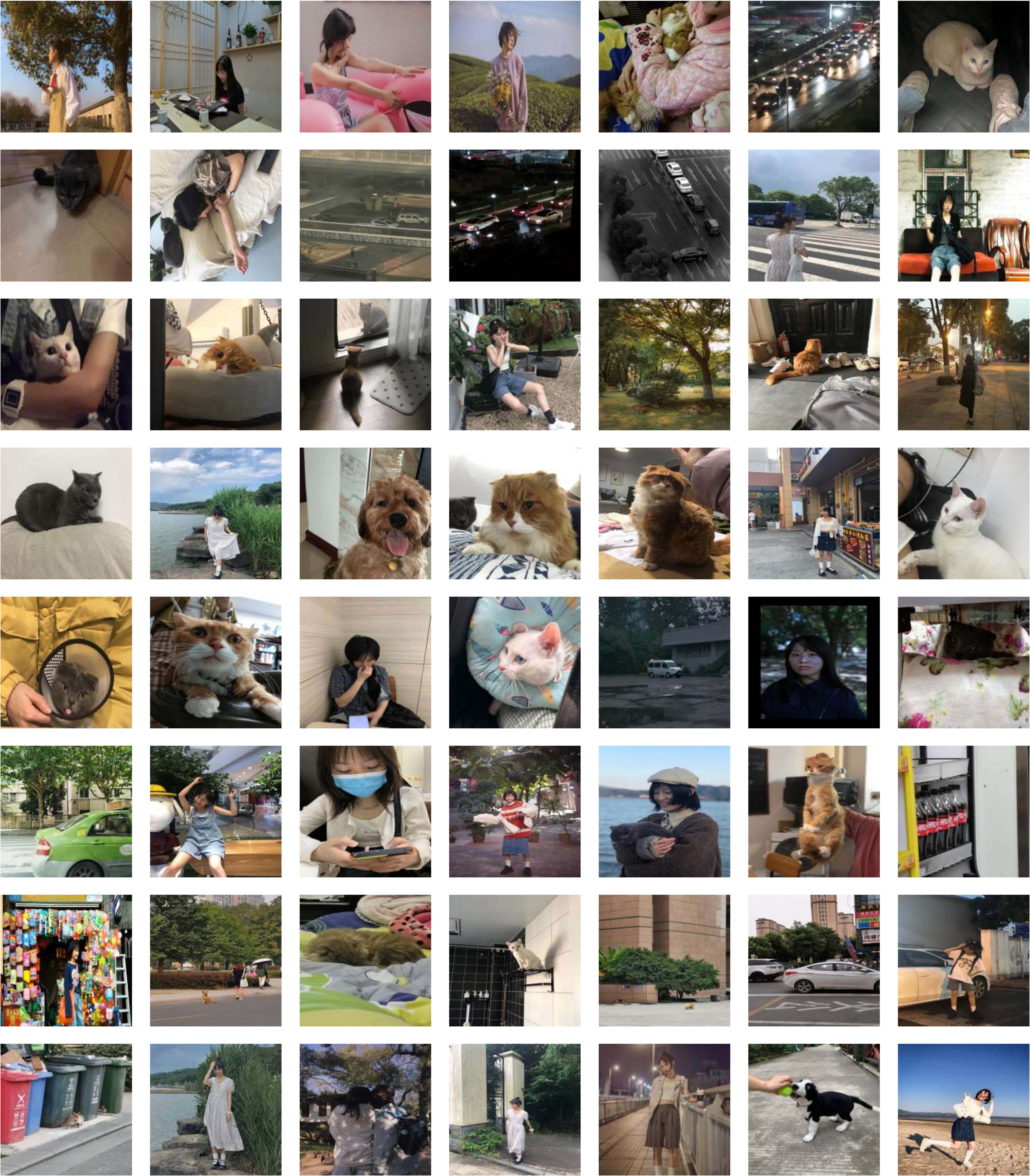} 
	\caption{Some samples from $ID11$.} \label{fig:ID11}
\end{figure*}

\begin{figure*}[t]
	\centering
	\includegraphics[width=.99\textwidth]{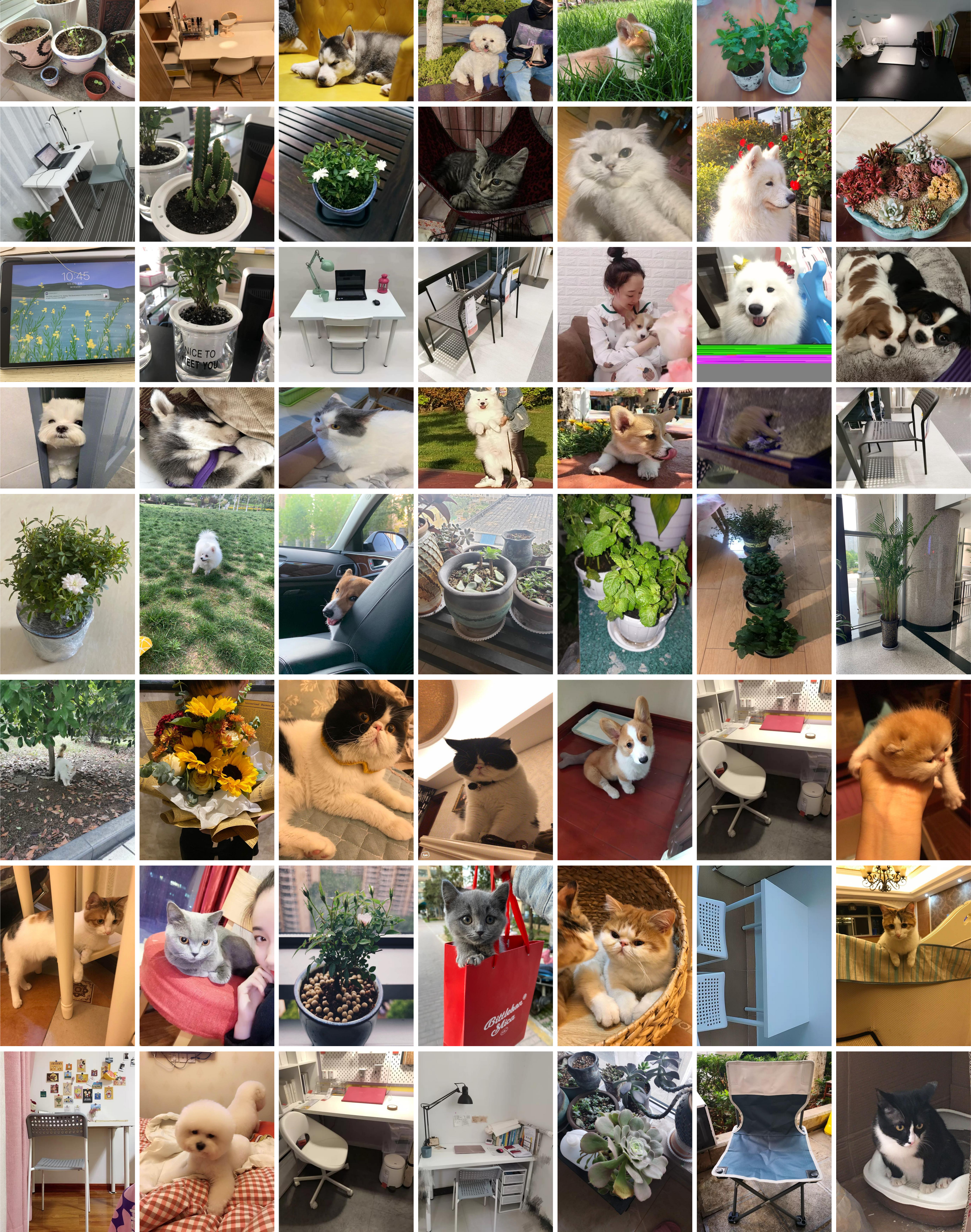} 
	\caption{Some samples from $ID12$.} \label{fig:ID12}
\end{figure*}

\begin{figure*}[t]
	\centering
	\includegraphics[width=\textwidth]{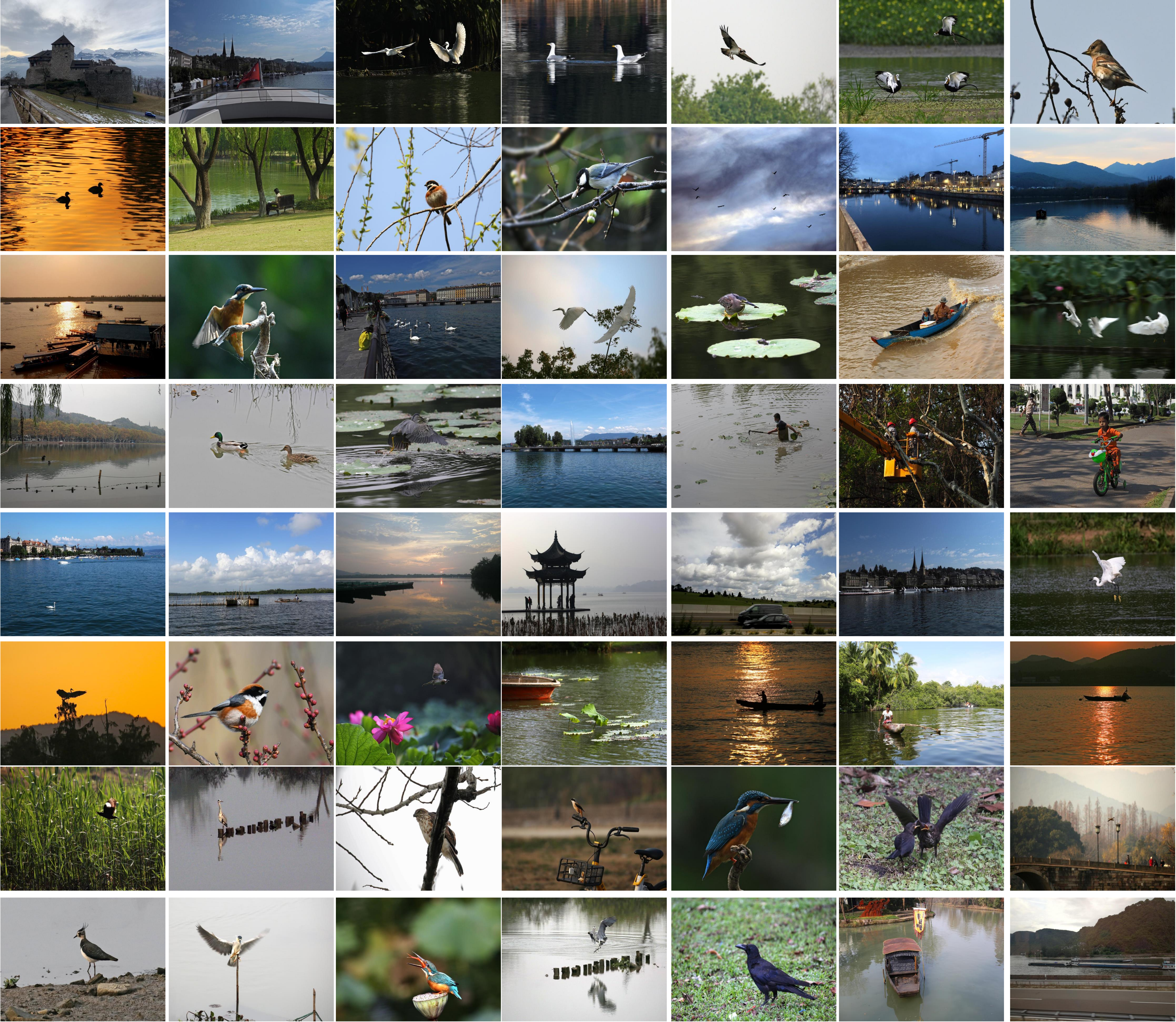} 
	\caption{Some samples from $ID13$.} \label{fig:ID13}
\end{figure*}

\begin{figure*}[t]
	\centering
	\includegraphics[width=.97\textwidth]{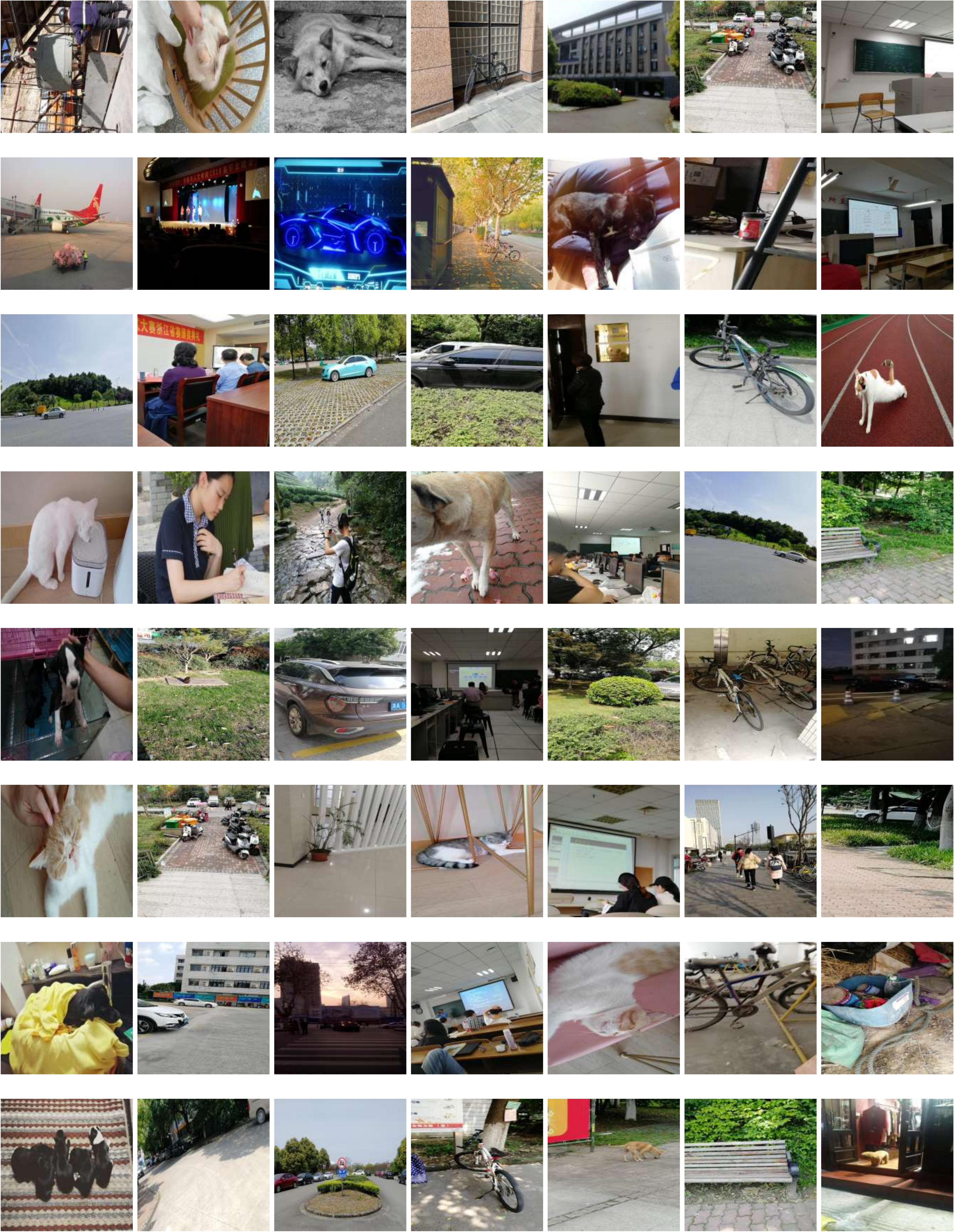} 
	\caption{Some samples from $ID14$.} \label{fig:ID14}
\end{figure*}

